\documentclass{article}

\PassOptionsToPackage{numbers,compress}{natbib}
\usepackage[preprint]{neurips_2026} 

\usepackage[utf8]{inputenc}
\usepackage[T1]{fontenc}
\usepackage{hyperref}
\usepackage{url}
\usepackage{booktabs}
\usepackage{amsfonts}
\usepackage{amsmath}
\usepackage{amssymb}
\usepackage{nicefrac}
\usepackage{microtype}
\usepackage{xcolor}
\usepackage{graphicx}
\usepackage{multirow}
\usepackage{xspace}
\usepackage{enumitem}
\usepackage{algorithm}
\usepackage{algorithmic}
\usepackage{natbib}
\usepackage{wrapfig}
\usepackage{pifont}
\usepackage{adjustbox}
\usepackage{graphicx}
\usepackage{makecell}
\usepackage{graphicx}
\usepackage{adjustbox}
\usepackage{graphicx}
\usepackage{subfig}
\usepackage{multirow}
\usepackage{amsmath}

\title{GHOST: Geometry-Hierarchical Online Streaming Token Eviction
  for Efficient 3D Reconstruction}

\author{
  Leyang Chen\thanks{Equal contribution},\enspace
  Junyi Wu\footnotemark[1],\enspace
  Zhiteng Li\footnotemark[1],\enspace
  Yulun Zhang\thanks{Corresponding author: Yulun Zhang, yulun100@gmail.com}\\
  Shanghai Jiao Tong University\\
}

\begin{document}

\maketitle

\begin{abstract}
Streaming 3D reconstruction from long monocular video sequences requires maintaining
a key-value~(KV) cache that grows linearly with sequence length, creating a severe
memory bottleneck.
Existing approaches either truncate the cache to a fixed set of anchor frames, leading
to reconstruction quality degradation, or rely on attention-score heuristics
that are agnostic to 3D scene structure,failing to preserve geometrically valuable
 tokens.
To address these problems, we present \textbf{GHOST} (\textbf{G}eometry-\textbf{H}ierarchical \textbf{O}nline
\textbf{S}treaming \textbf{T}oken Eviction), a training-free KV cache management
framework that exploits the model's own 3D geometry outputs to  evict
redundant tokens online.
GHOST introduces three mutually reinforcing innovations: a hierarchical dual-level
importance scoring scheme, a privilege mechanism that protects special
tokens from eviction, and a cosine-similarity-guided layer-wise budget allocation.
Experiments on various benchmarks show that GHOST preserves excellent reconstruction
quality while cutting the KV cache by nearly half and delivering $1.75\times$
faster inference compared to state-of-the-art methods.
Our code is available at https://github.com/lokiniuniu/GHOST.
\end{abstract}

\setlength{\abovedisplayskip}{2pt}
\setlength{\belowdisplayskip}{2pt}

\section{Introduction}
\label{sec:intro}

\begin{wrapfigure}{r}{0.44\columnwidth}
  \vspace{-4mm}
  \centering
  \includegraphics[width=0.44\columnwidth]{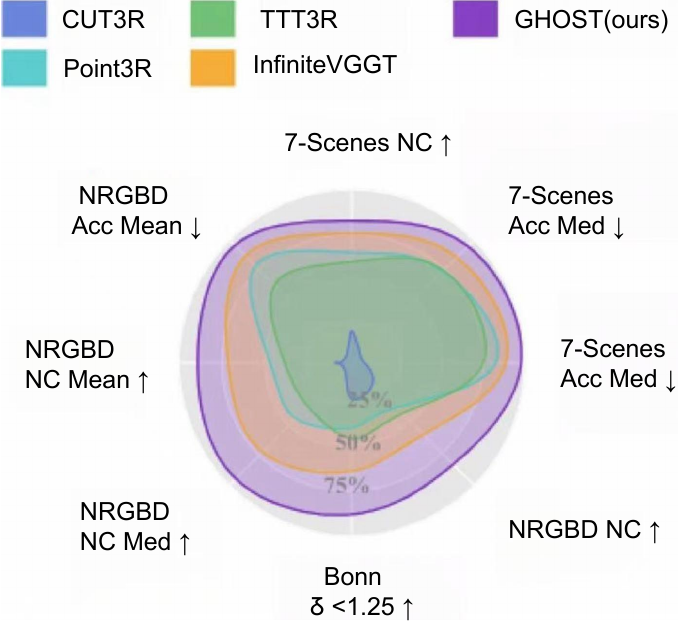}
  \vspace{-6mm}
  \caption{Radar comparison across 7-Scenes, NRGBD and Bonn (averaged over all input lengths; outer\,=\,better).
    GHOST consistently dominates all baselines on every axis.}
  \label{fig:radar}
  \vspace{-5mm}
\end{wrapfigure}

Transformer models~\cite{vaswani2017attention} have achieved remarkable results in 3D reconstruction
from monocular images~\cite{dust3r,mast3r,vggt}, learning to predict dense depth,
point maps, and camera poses in a single forward pass.
VGGT~\cite{vggt} extends this to multi-view sequences by attending jointly over all
input frames, delivering SOTA reconstruction quality.
However, joint attention scales quadratically with sequence length~\cite{zaheer2020big}, making it
infeasible for longer video-length inputs.

Enabling long-sequence 3D reconstruction~\cite{vggt,3dgs,balancegs} therefore requires a streaming paradigm where the model processes frames incrementally. StreamVGGT~\cite{streamvggt} pioneers
this direction by processing frames one at a time and maintaining a growing KV cache
of past observations.
InfiniteVGGT~\cite{infinitevggt} pushes this further by introducing an explicit token
eviction strategy: once the KV cache exceeds a fixed budget, tokens whose keys are
least similar to the current query are discarded  for new frames.
Yet this strategy has a fundamental limitation, as attention scores from the
current query frame cannot faithfully reflect the long-term geometric value of
historical tokens~\cite{li2025analyzing}.

\begin{wrapfigure}{r}{0.55\columnwidth}
  \centering
  \begin{minipage}{0.48\linewidth}
    \centering
    \includegraphics[width=\linewidth]{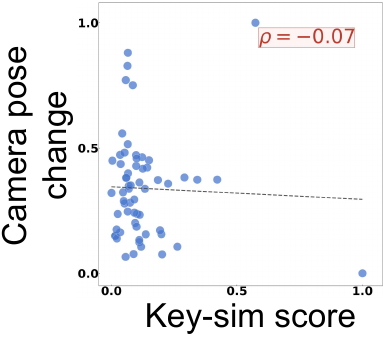}
    \hspace{8mm}
  \end{minipage}
  \hfill
  \begin{minipage}{0.48\linewidth}
    \centering
    \includegraphics[width=\linewidth]{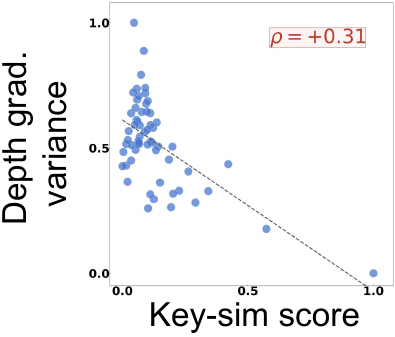}
    \hspace{8mm}
  \end{minipage}
  \vspace{-5mm}
  \caption{Correlation between Key-sim score and two frame attributes: 
  Left: Negligible linear correlation between Key-sim score and camera pose change ($\rho=-0.07$); 
  right: Moderate positive linear correlation between Key-sim score and depth gradient variance ($\rho=+0.31$). 
  Dashed lines denote linear fitting trends.}
  \label{fig:obs1}
  \vspace{-15pt}
\end{wrapfigure}
As shown in Figure~\ref{fig:obs1}, the Spearman correlation between key-similarity
scores and camera pose change is only $-0.07$, and with depth gradient variance
$-0.31$. It confirms that key-similarity cannot make full use of  the signals that determine a token's long-term geometric value.
However, both signals are already available  from the model's own outputs ~\cite{vggt}.
Moreover, transformer layers vary significantly in transformation strength: blocks
with near-identical input and output representations can tolerate smaller token
budgets without accuracy loss~\cite{yang2024kvsharer}, yet uniform allocation
ignores this heterogeneity.

Building on these observations, we propose \textbf{GHOST}
(\textbf{G}eometry-\textbf{H}ierarchical \textbf{O}nline \textbf{S}treaming
\textbf{T}oken Eviction), a training-free 
eviction module for StreamVGGT~\cite{streamvggt}.
GHOST scores each cached token by a \emph{hierarchical dual-level importance}
signal derived entirely from the model's own outputs: a \emph{frame-level}
component integrating camera pose change, depth gradient variance, and temporal
recency,which captures whether a viewpoint is geometrically distinctive; a
\emph{token-level} component integrating visual saliency, depth confidence, and
3D point confidence, which identifies which patches carry reliable reconstruction value.
The two levels are combined per-patch and updated incrementally online, incurring
negligible overhead.
A \emph{special-token privilege mechanism} deterministically boosts camera and
register tokens above all patch tokens, preventing catastrophic eviction of
globally critical pose and structure encodings.
A \emph{cosine-similarity-guided layer-wise budget allocation} concentrates the
global token budget where transformer layers perform the strongest transformations,
while reducing budgets for near-identity later layers.

Our contributions are fourfold:
\vspace{-1mm}
\begin{itemize}
\vspace{-1mm}
\item \textbf{Hierarchical dual-level importance scoring}: we decompose token
importance into geometry-aware frame-level and token-level components derived from
the model's own outputs, updated incrementally online without extra forward passes.

\vspace{-1mm}
\item \textbf{Special-token privilege mechanism}: we introduce a
deterministic importance boost that protects camera and register tokens from
eviction, preventing corruption of global pose and structure encodings.

\vspace{-1mm}
\item \textbf{Cosine-similarity-guided layer-wise budget allocation}: we
profile transformer layers offline and concentrate the token budget where
transformations are strongest.

\vspace{-1mm}
\item \textbf{Comprehensive empirical evaluation}: as shown in Fig~\ref{fig:radar}, experiments on
7-Scenes~\cite{sevenscenes}, NRGBD~\cite{nrgbd}, Bonn~\cite{bonn}, and Long3D~\cite{infinitevggt} across four input-length regimes show that GHOST
 outperforms near all baselines at matched budgets.GHOST also maintains competitive
accuracy while cutting the KV cache by nearly half and delivering $1.75\times$
faster inference compared to state-of-the-art methods.
\end{itemize}

\vspace{-3mm}
\section{Related Work}
\label{sec:related}
\vspace{-3mm}
\noindent\textbf{Streaming and long-sequence 3D reconstruction.}
Feed-forward multi-view reconstruction models such as DUSt3R~\citep{dust3r} and
MASt3R~\citep{mast3r} regress 3D point maps and camera poses from image pairs,
enabling downstream dense reconstruction by global alignment.
VGGT~\citep{vggt} scales to multi-view sequences by processing all frames jointly
with a full cross-frame attention transformer.
StreamVGGT~\citep{streamvggt} adopts a streaming inference paradigm where each new
frame attends to a fixed-size KV cache populated by prior frames, enabling causal
online processing.
InfiniteVGGT~\citep{infinitevggt} extends this to unbounded sequences by evicting
tokens ranked by key-space cosine similarity to the current query.
Our work proposes an alternative eviction criterion based on multi-modal 3D geometry
signals, which we show to be more informative than attention-based rankings.

\begin{figure*}[t]
    \centering
    \begin{minipage}[b]{0.93\textwidth}
        \centering
        \raisebox{0.9\height}{\rotatebox{90}{\normalsize RGB}}
        \hspace{1mm}
        \includegraphics[width=0.228\textwidth]{ 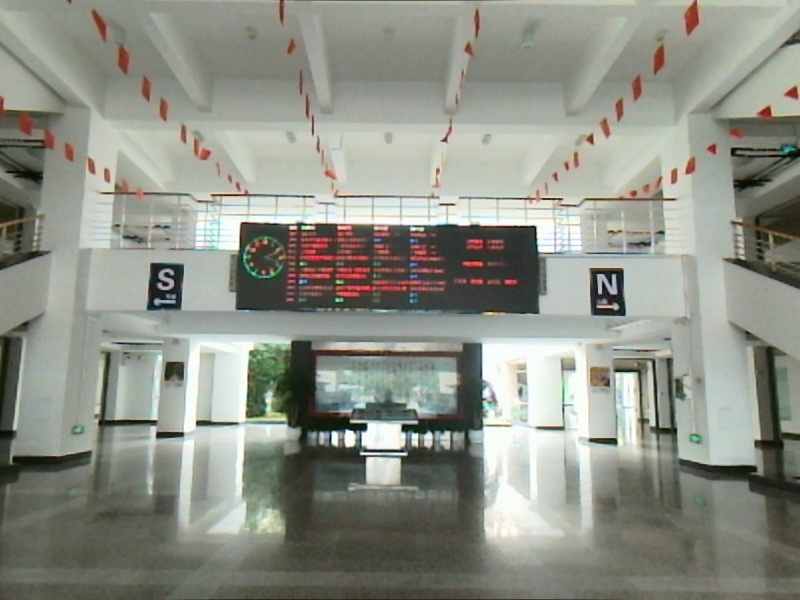}
        \includegraphics[width=0.228\textwidth]{ 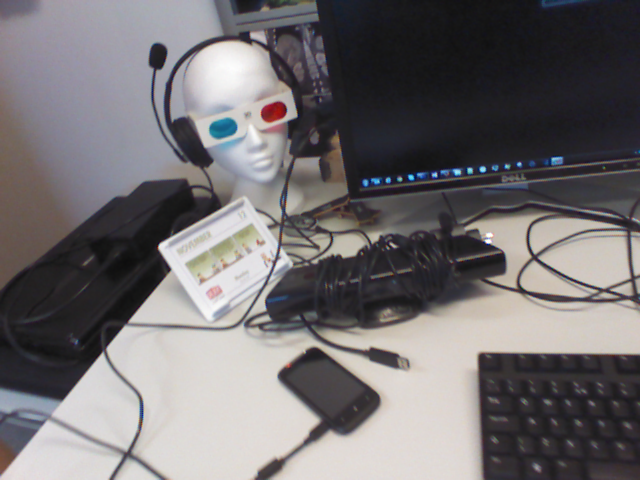}
        \includegraphics[width=0.228\textwidth]{ 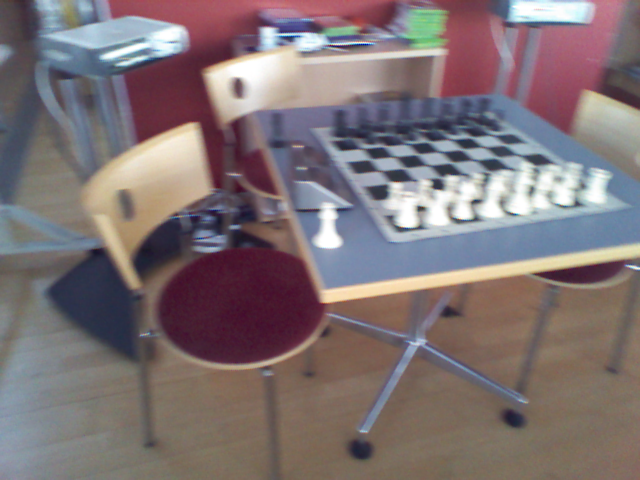}
        \includegraphics[width=0.228\textwidth]{ 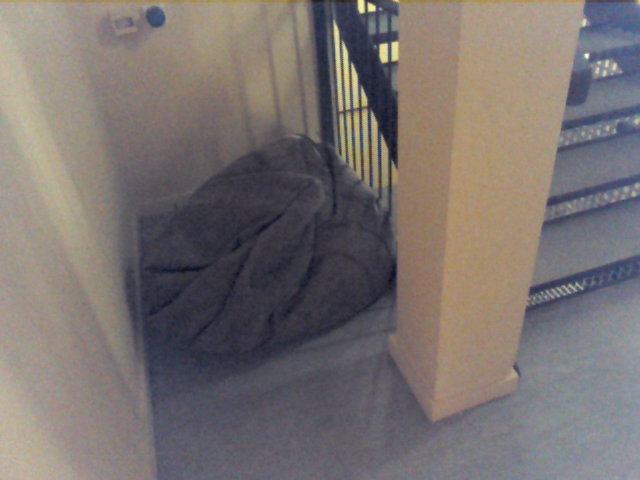} \\[6pt]
        \vspace{-1.5mm}
        \raisebox{0.8\height}{\rotatebox{90}{\large Score}}
        \hspace{0.5mm}
        \includegraphics[width=0.228\textwidth]{ 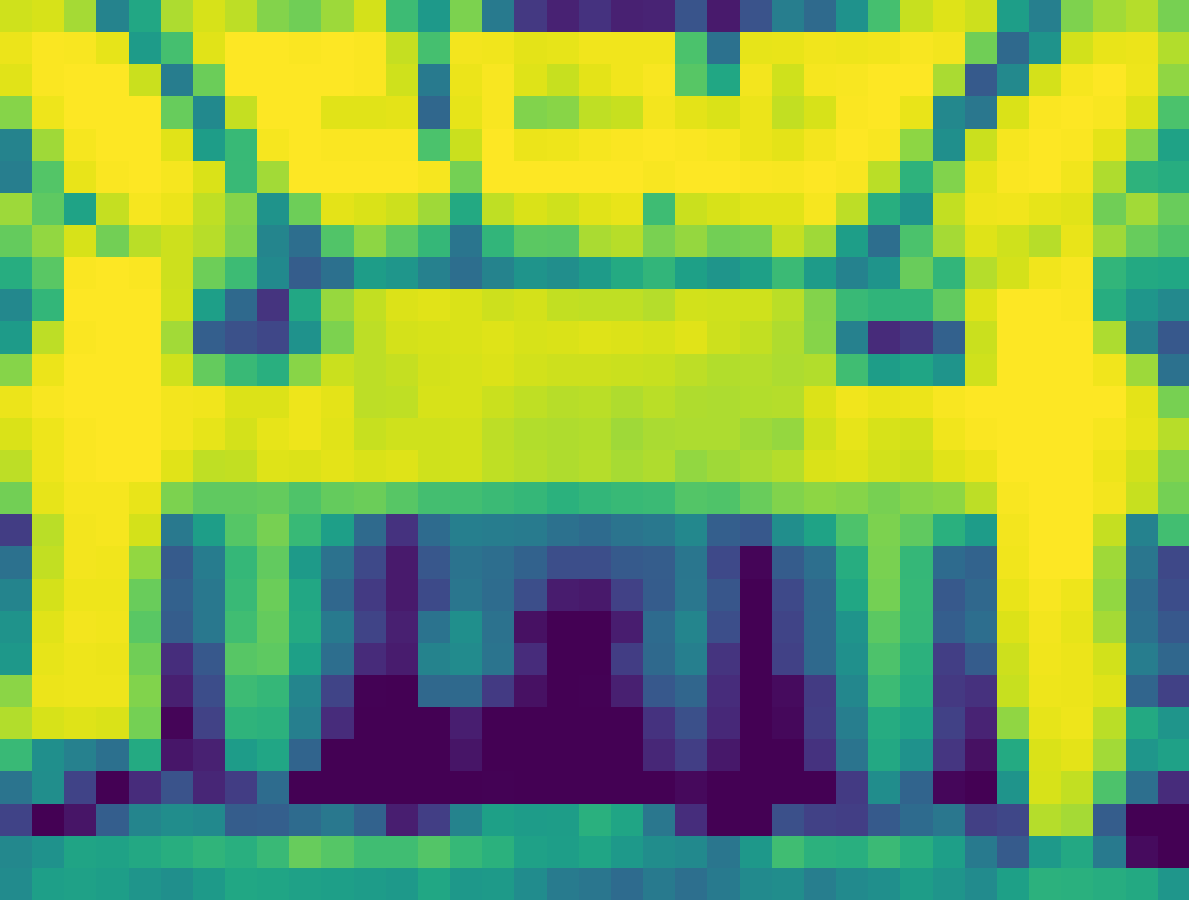}
        \includegraphics[width=0.228\textwidth]{ 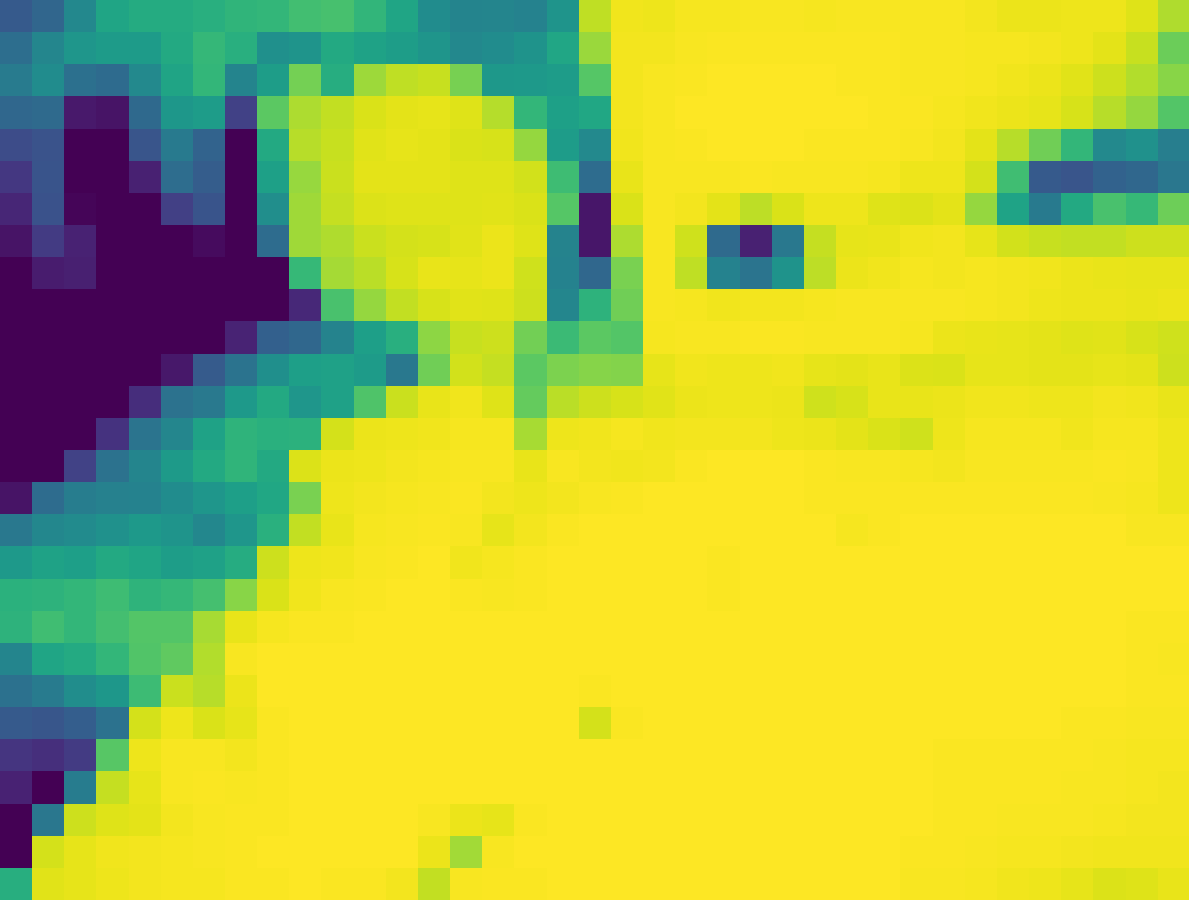}
        \includegraphics[width=0.228\textwidth]{ 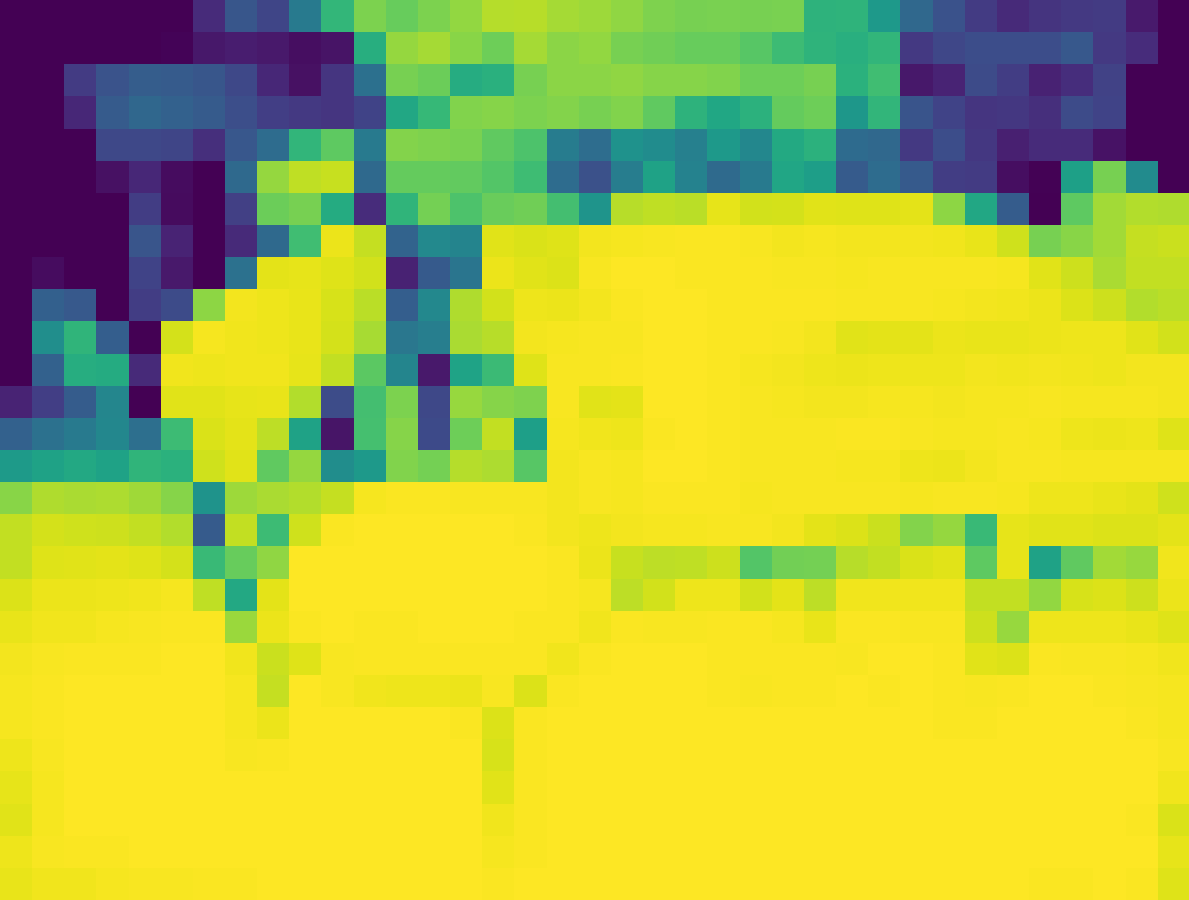}
        \includegraphics[width=0.228\textwidth]{ 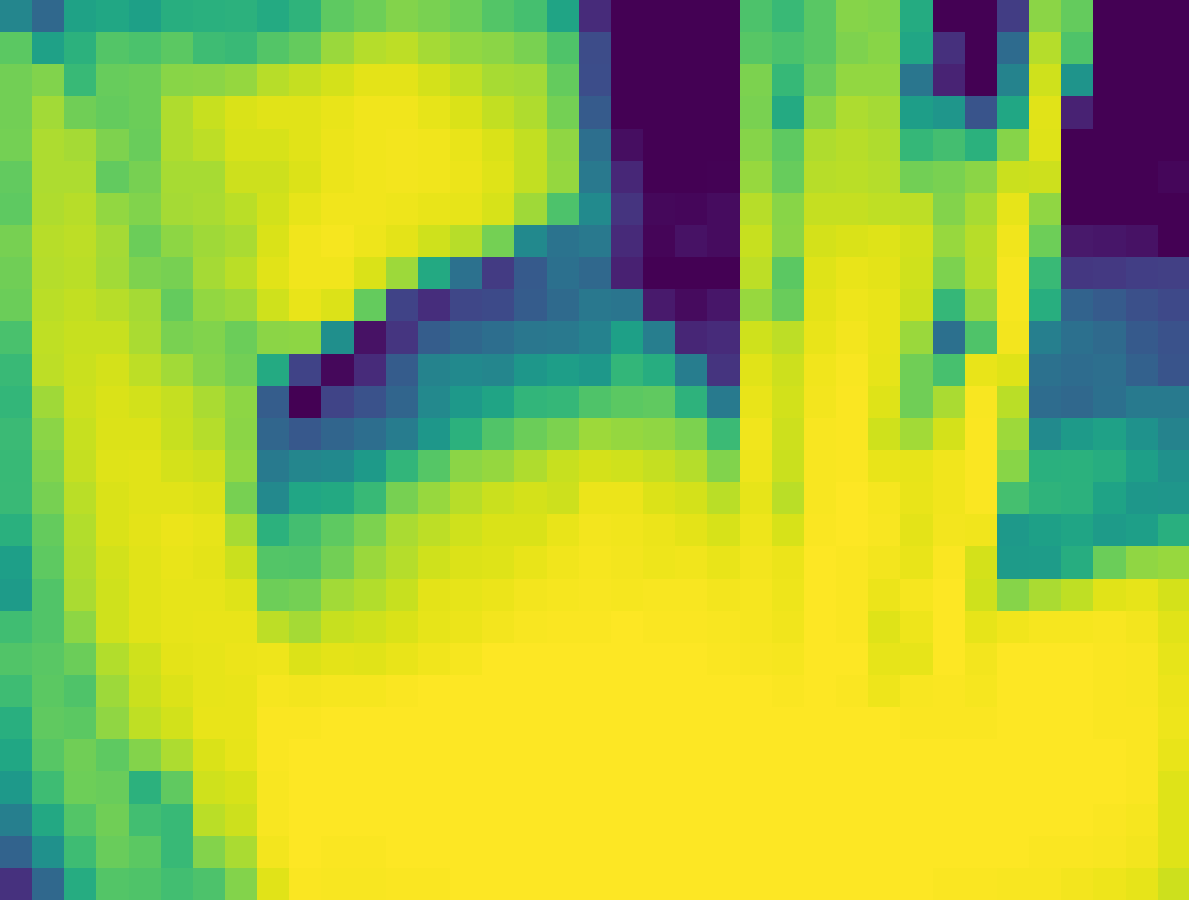}
    \end{minipage}
    \hspace{-4mm}
    \begin{minipage}[b]{0.06\textwidth}
        \centering
        \includegraphics[height=0.1\textheight]{ 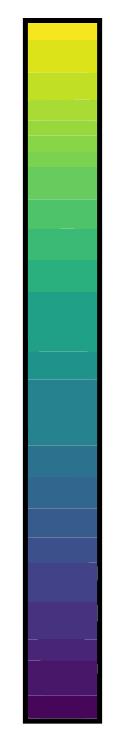}
    \end{minipage}
    \vspace{-2mm}
    \caption{Top row: Raw RGB inputs from Long3D Lecture Hall, 7-Scenes Heads, 7-Scenes Chess, and 7-Scenes Stairs.
Bottom row: Corresponding GHOST token importance heatmaps, illustrating that GHOST prioritizes visually/structurally salient regions across diverse scenarios.}
    \label{fig:ghost_results}
    \vspace{-5.5mm}
\end{figure*}

\vspace{-1mm}
\noindent\textbf{KV cache compression for transformers.}
The memory bottleneck of long-context transformers~\cite{streamingvlm,quantcache} has spurred extensive research on
KV cache compression~\cite{streamingllm,h2o,snapkv,xu2026specontext,wu2025flashedit}.
StreamingLLM~\citep{streamingllm} retains only ``attention sink'' tokens and recent
frames, discarding most of historical context.
H2O~\citep{h2o} identifies ``heavy hitter'' tokens via accumulated attention scores
and preserves them across steps.
SnapKV~\citep{snapkv} clusters KV entries by key similarity to select a representative
set.
PyramidKV~\citep{pyramidkv} allocates larger budgets to lower transformer layers,
motivated by the observation that lower layers attend more broadly.
All of these methods derive importance from attention patterns or token similarities
\emph{within the language modality}.
Our work departs fundamentally by exploiting \emph{3D geometry outputs}---depth
confidence, point confidence, and camera pose, as the primary importance signal,
a direction that has not been explored in prior KV cache compression work.

\vspace{-1mm}
\noindent\textbf{Token reduction in vision transformers.}
EViT~\citep{feng2024evit} and DynamicViT~\citep{rao2021dynamicvit} progressively discard tokens
based on attention-derived relevance scores.
A-ViT~\citep{yin2022vit} learns per-token adaptive halting thresholds, while
Evo-ViT~\citep{xu2022evo} routes tokens into slow and fast processing streams.
On the merging side, ToMe~\citep{bolya2022token} fuses redundant tokens via bipartite
matching of key-space similarities, and DiffRate~\citep{chen2023diffrate} relaxes the
compression rate into a differentiable objective for joint optimisation.
or video transformers, STTS~\citep{wang2022efficient} jointly selects tokens along spatial
and temporal axes to exploit inter-frame redundancy.
Evict3R~\cite{mahdi2025evict3r} performs training-free token eviction for memory-bounded streaming 3D reconstruction, selecting tokens via attention-based importance and KV-cache budgeting
None of these methods generalise to the streaming 3D reconstruction setting,
where token importance is temporally structured and tied to geometric
distinctiveness; our scoring instead exploits depth, point confidence, and
camera-pose signals produced by the model itself.

\vspace{-3mm}
\section{Motivation}
\label{sec:motivation}
\vspace{-2mm}
\subsection{Observation and Motivation}
\vspace{-2mm}
\noindent\textbf{Key-similarity eviction lacks geometric grounding.}
InfiniteVGGT~\citep{infinitevggt} retains tokens whose keys are least similar to the others.
As shown in Figure~\ref{fig:obs1}, however, Key-sim scores correlate negligibly with camera pose change ($\rho{=}{-}0.07$) and moderately with depth gradient variance ($\rho{=}{-}0.31$).
Since higher Key-sim tokens are retained, the preserved  are moderately uncorrelated with geometric distinctiveness.Thus frames with large camera shifts or rich structural detail cannot be fully preserved.

\vspace{-1mm}
\noindent\textbf{Importance is inherently two-level.}
Figure~\ref{fig:ghost_results} shows that within a single frame, structurally salient patches carry far more reconstruction value than flat, textureless regions.
Token importance is thus two-level: a \emph{frame-level} component captures whether a viewpoint is geometrically distinctive, while a \emph{token-level} component identifies which patches within that frame are most informative.

\vspace{-1mm}
\noindent\textbf{Transformer layers exhibit varying transformation strengths.}
Transformer layers differ substantially in how strongly they transform their inputs~\citep{yang2024kvsharer}.
Near-identity layers tolerate smaller token budgets without accuracy loss, whereas early high-transformation layers benefit from larger budgets.
Uniform allocation ignores this heterogeneity, motivating a layer-wise budget strategy guided by each layer's transformation strength.
Specifically, we allocate larger budgets to layers with stronger transformations and smaller budgets to near-identity layers.
\begin{figure}[t]
  \centering
 \includegraphics[width=\linewidth]{ 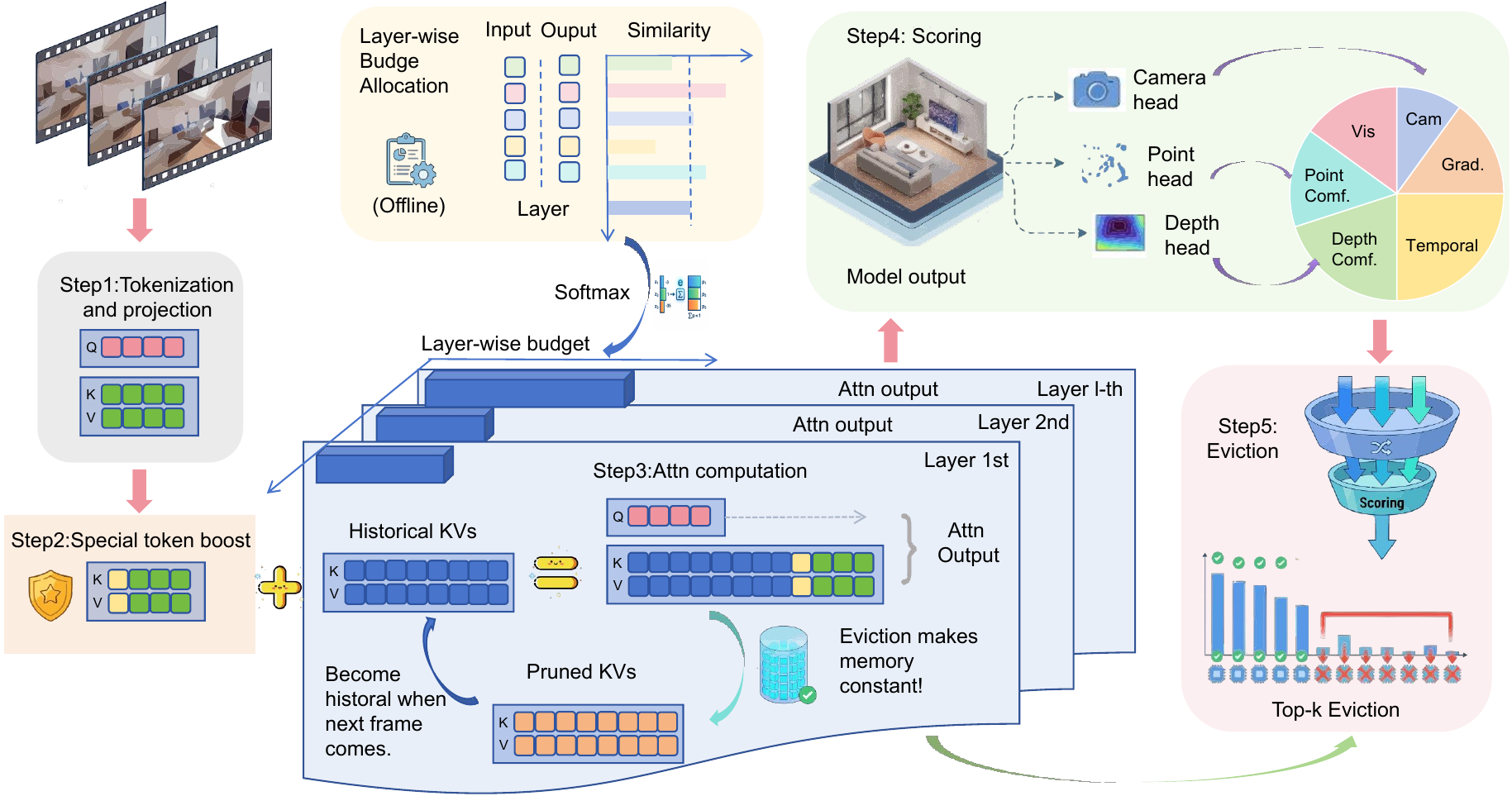}
 \vspace{-4.5mm}
  \caption{GHOST inference pipeline.
    \textbf{Offline:} Cosine-similarity profiling allocates per-layer budgets.
    \textbf{Online:} An eviction mode that prunes KV cache to  layer-wise budget computed offline with Geometry-Hierarchical Importance scoring and Special token boost .}
    \vspace{-5.5mm}
  \label{fig:overview}
\end{figure}

\vspace{-3mm}
\section{GHOST: Geometry-Hierarchical Online Streaming Token Eviction}
\label{sec:method}
\vspace{-2mm}
\subsection{Preliminaries}
\label{sec:prelim}
\vspace{-2mm}

StreamVGGT~\cite{streamvggt} processes frames $\{I_t\}_{t=1}^{T}$ with $N_p = H_p \times W_p$ patch tokens,
prepending $p_0 = 1{+}R$ special tokens ($\mathbf{c}_t$ and $\{\mathbf{r}_t^i\}_{i=1}^{R}$) for a total of $N$ tokens per frame.
$L$ global self-attention layers maintain per-layer KV caches with total budget $B_{\text{total}}$.
Outputs comprise depth $\{d_t, c^d_t\}$, point maps $\{P_t, c^p_t\}$, and pose $\boldsymbol{\pi}_t = (\mathbf{T}_t, \mathbf{q}_t, f_t)$.
Our final goal is to allocate the limited KV cache budget across layers to maximize overall prediction accuracy.

GHOST assigns per-patch importance $\phi(t, p) = w_f s_{\text{frame}}(t) + w_k s_{\text{token}}(t, p)$,
where $s_{\text{frame}}$ combines camera motion $s_{\text{cam}}$, depth variance $s_{\text{geo}}$, and recency $s_{\text{temp}}$,
while $s_{\text{token}}$ aggregates visual saliency $s_{\text{sal}}$ from feature map $\mathbf{F}_t$, depth confidence $c^d_t(p)$, and point confidence $c^p_t(p)$.
Special tokens receive a deterministic boost with hyperparameters $\Delta_{\text{boost}}, \epsilon_{\text{tb}}, r_{p_{\text{sp}}}$.
Per-layer budgets $B_\ell$ are allocated by importance $\pi_\ell \propto \exp(a_\ell/\tau)$ where $a_\ell = 1 - \bar{\rho}_\ell$
is derived from offline cosine similarity statistics.Figure~\ref{fig:overview} provides an overview of the GHOST inference pipeline.

\vspace{-3mm}
\subsection{Hierarchical Dual-Level Importance Scoring}
\label{sec:importance}
\vspace{-2mm}
GHOST decomposes token importance into a frame-level component $s_{\text{frame}}(t)$
and a token-level component $s_{\text{token}}(t, p)$, combined into a per-patch
importance score $\phi(t, p)$.
This two-level decomposition mirrors the hierarchical structure of 3D scene understanding:
relevant information must first be identified at the viewpoint level (which frames are
geometrically distinctive?) and then refined at the patch level (which spatial regions
within those frames are most informative?).

\vspace{-2mm}
\subsubsection{Frame-Level Importance}
\vspace{-1mm}
The frame-level score $s_{\text{frame}}(t)$ captures the global geometric value of an
entire frame as a KV-cache contributor, reflecting distinct aspects of a frame's
reconstruction value.

\vspace{-1mm}
\noindent\textbf{Camera pose change.}
Frames from significantly different viewpoints provide complementary geometric coverage
that cannot be inferred from nearby frames.
Redundant frames captured from nearly identical poses duplicate information already
present in the cache, whereas frames with large pose differences introduce new scene
geometry and viewing directions critical for triangulation and novel-view synthesis.
Let $\boldsymbol{\pi}_t = (\mathbf{T}_t, \mathbf{q}_t, f_t)$ encode translation,
unit quaternion, and focal length.
The camera change score relative to the preceding frame is:
\begin{equation}
  s_{\text{cam}}(t) \;=\;
    \|\mathbf{T}_t - \mathbf{T}_{t-1}\|_2 + 1 - |\mathbf{q}/\|\mathbf{q}\|_t \cdot \mathbf{q}/\|\mathbf{q}\|_{t-1}|.
  \label{eq:cam}
\end{equation}
The two terms respectively penalise translational proximity and rotational similarity,
ensuring that both lateral movement and pure rotations are treated as geometrically
meaningful changes.

\vspace{-1mm}
\noindent\textbf{Depth gradient variance.}
Frames with high spatial depth variation contain richer geometric detail
(edges, object boundaries, depth discontinuities) and are harder to approximate by
neighbouring frames.
Conversely, frames depicting large uniform surfaces---such as blank walls or open
floors---contribute little unique structural information and are less critical to retain.
Given depth map $d_t$:
\begin{equation}
  s_{\text{geo}}(t) \;=\; \operatorname{Var}\!\bigl(\|\nabla d_t\|\bigr).
  \label{eq:geo}
\end{equation}
High variance in the depth gradient magnitude indicates a spatially diverse scene
structure, making the frame a geometrically rich reference for reconstruction.
Such frames are therefore more informative for preserving fine-grained 3D structure across time.
Accordingly, we prioritize them during memory retention or budget allocation to improve reconstruction fidelity.

\noindent\textbf{Temporal recency.}
Under causal streaming, the model processes frames sequentially and must infer scene
state incrementally.
Recent frames share greater contextual overlap with the current query frame---in terms
of scene content, lighting, and object configuration---and therefore provide more
directly applicable geometric context.
We assign $s_{\text{temp}}(t) = t / T_{\text{cur}}$,
where $T_{\text{cur}}$ is the current frame index.
This linear schedule prioritises recency without entirely discarding older frames,
which may still encode unique viewpoints not revisited since.

All three raw scores operate on different scales and distributions, so each is mapped
through sigmoid $\sigma(\cdot)$ to $[0,1]$ before aggregation.
The frame-level score is then their normalised weighted sum:
\begin{equation}
  s_{\text{frame}}(t)
    \;=\;
    \frac{w_{\text{cam}}\,\sigma(s_{\text{cam}}(t))
         + w_{\text{geo}}\,\sigma(s_{\text{geo}}(t))
         + w_{\text{temp}}\,\sigma(s_{\text{temp}}(t))}
         {\max_{t' \in \mathcal{T}}\bigl(w_{\text{cam}}\,\sigma(s_{\text{cam}}(t'))
         + w_{\text{geo}}\,\sigma(s_{\text{geo}}(t'))
         + w_{\text{temp}}\,\sigma(s_{\text{temp}}(t'))\bigr) + \epsilon},
  \label{eq:frame}
\end{equation}
where $t' \in \mathcal{T}$ ranges over all frames currently stored in the KV cache.
Division by the maximum value across all cached frames ensures $s_{\text{frame}} \in [0,1]$, thus
keeping it commensurate with the token-level score in the final combination.
This normalisation makes the scoring robust to variations in scene dynamics and sequence length.
It also enables consistent weighting across different video segments without requiring per-sequence tuning.
Consequently, $s_{\text{frame}}(t)$ can be directly integrated with token-level importance for unified budget allocation.

\vspace{-3mm}
\subsubsection{Token-Level Importance}
\vspace{-2mm}
While the frame-level score determines how valuable a viewpoint is as a whole,
the token-level score $s_{\text{token}}(t, p)$ provides fine-grained discrimination
\emph{within} a frame by measuring the geometric informativeness of each individual patch.
Even within a highly distinctive frame, only a subset of patches---those at structural
boundaries or exhibiting high reconstruction confidence---contribute disproportionately
to scene understanding.

\noindent\textbf{Visual saliency.}
High-gradient patches correspond to edges, corners, and texture-rich regions that anchor
geometric correspondence and are thus most informative for depth and point estimation.
Uniform or featureless patches, by contrast, provide weak cues and are more readily
interpolated from neighbouring tokens.
Let $\mathbf{F}_t \in \mathbb{R}^{H_p \times W_p \times d}$ be the patch feature map
extracted from the visual encoder.
Saliency is measured as the spatial gradient magnitude of the feature representation:
\begin{equation}
  s_{\text{sal}}(t, p)
    \;=\;
    \sqrt{\bigl\|\delta_x \mathbf{F}_t(p)\bigr\|^2 + \bigl\|\delta_y \mathbf{F}_t(p)\bigr\|^2}\,,
  \label{eq:sal}
\end{equation}
where $\delta_x, \delta_y$ denote horizontal and vertical finite differences over the
spatial patch grid.
Computing saliency in feature space rather than raw pixel space makes it robust to
illumination variation while remaining sensitive to semantically meaningful structure.

\noindent\textbf{Depth and point confidence.}
Model-predicted confidence scores provide a direct signal of which patches yield
reliable 3D estimates.
The depth head outputs a per-patch confidence $c^d_t(p)\in[0,1]$ and the point head
outputs $c^p_t(p)\in[0,1]$; both reflect the model's epistemic certainty about the
geometric reconstruction at patch $p$.
Patches with low confidence---typically arising from occlusion boundaries,
reflective surfaces, or textureless regions---contribute noisier estimates and are
therefore assigned lower importance.
For past frames, confidence maps are pooled to patch resolution via adaptive average
pooling and stored as lightweight per-patch metadata alongside the KV cache.
The token-level score aggregates all three signals:
\begin{equation}
  s_{\text{token}}(t, p)
    \;=\;
    \frac{w_{\text{sal}}\,\sigma(s_{\text{sal}}(t,p))
         + w_{\text{dc}}\,\sigma(c^d_t(p))
         + w_{\text{pc}}\,\sigma(c^p_t(p))}
         {\max_{t' \in \mathcal{T},\, p' \in \mathcal{P}}\bigl(
           w_{\text{sal}}\,\sigma(s_{\text{sal}}(t',p'))
         + w_{\text{dc}}\,\sigma(c^d_{t'}(p'))
         + w_{\text{pc}}\,\sigma(c^p_{t'}(p'))\bigr) + \epsilon},
  \label{eq:token}
\end{equation}
where $t' \in \mathcal{T}$ iterates over all cached frames and
$p' \in \mathcal{P}$ iterates over all $N_p$ patch positions within each frame.
The joint maximum over $(t', p')$ normalises scores globally across the entire
cache so that $s_{\text{token}} \in [0,1]$ and patch scores from different frames
remain directly comparable during eviction.

\vspace{-2mm}\vspace{-2mm}
\subsubsection{Combined Score}
\vspace{-2mm}
As established in Section~\ref{sec:motivation}, token importance is inherently \emph{two-level}: a frame-level component captures whether a viewpoint is geometrically distinctive (large camera displacement, rich depth structure, high recency), while a token-level component identifies which patches \emph{within} that frame carry the most reconstruction value (salient edges, reliable depth and point estimates).
Neither level alone is sufficient.
A frame-level score applied uniformly across all patches of a highly distinctive viewpoint would waste budget on flat, textureless regions of that frame.
Conversely, a token-level score that ignores inter-frame geometry may retain visually active patches from redundant viewpoints while discarding tokens from geometrically critical but low-contrast frames.

The combined per-patch importance score integrates both levels via a weighted sum:
\begin{equation}
  \phi(t, p) \;=\; w_f\,s_{\text{frame}}(t) + w_k\,s_{\text{token}}(t, p),
  \label{eq:combined}
\end{equation}
where $w_f + w_k = 1$, $w_f, w_k \geq 0$.
The additive structure is deliberate: $s_{\text{frame}}(t)$ acts as a \emph{prior} that uniformly elevates or suppresses all patches from frame $t$ according to its global geometric value, while $s_{\text{token}}(t,p)$ provides \emph{within-frame discrimination} by further differentiating patches inside the same frame.
A patch therefore receives a high combined score only when it belongs to a geometrically important viewpoint \emph{and} constitutes an informative region within that viewpoint, naturally capturing the two-level structure of token importance identified in our observations.
The final score is re-normalised to $[0,1]$ across all candidate tokens before eviction decisions are made.

\begin{table}[t!]
  \centering
  \footnotesize
  \caption{Quantitative comparison on Bonn under different input lengths.
    \textbf{Bold}: best result in each column per input-length block.}
  \label{tab:main_bonn}
  \small
  \renewcommand{\arraystretch}{0.88}
  \begin{tabular}{lcccc}
    \toprule
    
      & \multicolumn{2}{c}{Input 200 / 300}
      & \multicolumn{2}{c}{Input 400 / 500} \\
    \cmidrule(lr){2-3}\cmidrule(lr){4-5}
      \multirow{-2}{*}{Method}& Abs Rel$\downarrow$ & $\delta < 1.25\uparrow$
      & Abs Rel$\downarrow$ & $\delta < 1.25\uparrow$ \\
    \midrule
    VGGT (\textit{Off.})~\cite{vggt}     & \textit{OOM} / \textit{OOM} & \textit{OOM} / \textit{OOM} & \textit{OOM} / \textit{OOM} & \textit{OOM} / \textit{OOM} \\
    StreamVGGT~\cite{streamvggt}         & \textit{OOM} / \textit{OOM} & \textit{OOM} / \textit{OOM} & \textit{OOM} / \textit{OOM} & \textit{OOM} / \textit{OOM} \\
    CUT3R~\cite{cut3r}                   & 0.072 / 0.089 & 0.947 / 0.938 & 0.090 / 0.084 & 0.934 / 0.939 \\
    Point3R~\cite{point3r}               & 0.069 / 0.081 & 0.954 / 0.946 & 0.081 / 0.081 & 0.945 / 0.946 \\
    TTT3R~\cite{ttt3r}                   & 0.068 / 0.079 & 0.953 / 0.949 & 0.078 / 0.076 & 0.951 / 0.953 \\
    InfiniteVGGT~\cite{infinitevggt}     & 0.063 / 0.072 & 0.964 / 0.958 & 0.070 / 0.069 & 0.958 / 0.960 \\
    GHOST (ours)                         & \textbf{0.054} / \textbf{0.062} & \textbf{0.971} / \textbf{0.970} & \textbf{0.064} / \textbf{0.061} & \textbf{0.962} / \textbf{0.964} \\
    \bottomrule
  \end{tabular}
  \vspace{-5mm}
\end{table}

\vspace{-2mm}
\subsection{Special-Token Privilege Mechanism}
\label{sec:special}
\vspace{-2mm}

\begin{wrapfigure}{b}{0.46\textwidth} 
\vspace{-11.3mm}
  \centering
  \includegraphics[width=\linewidth]{ 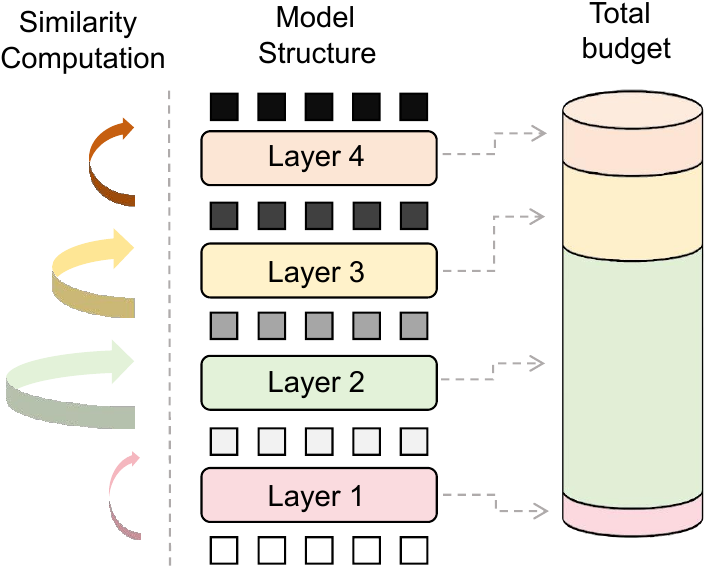}  
  \vspace{-4mm}
  \caption{Layer-wise budget allocation guided by cosine similarity.
    Larger input--output colour discrepancy and larger arrows indicate lower
    $\bar{\rho}_\ell$; the cylinder shows how
    $B_{\text{total}}$ is distributed, with such layers receiving larger $B_\ell$.}
  \label{fig:similarity}  
\end{wrapfigure}

Camera tokens $\mathbf{c}_t$ and register tokens $\{\mathbf{r}_t^i\}$ encode global
scene geometry state and structural priors.
Evicting these tokens can corrupt pose estimation and scene
globalisation, yet standard importance scoring may rank them below informative patch
tokens due to their low visual saliency.
We introduce a \emph{deterministic importance boost} for special tokens:
\begin{equation}
  \phi(t, p_{\mathrm{sp}})
    \;=\; s_{\text{frame}}(t) + \Delta_{\text{boost}} + \epsilon_{\text{tb}}\cdot r_{p_{\mathrm{sp}}},
  \label{eq:boost}
\end{equation}
where $\Delta_{\text{boost}}$ ensures all special tokens score above any patch
token (since both $s_{\text{frame}}$ and $\phi$ are in $[0,1]$ for patches), and
$r_{p_{\mathrm{sp}}} \in \{0, 1, \ldots, p_0{-}1\}$ is the intra-frame rank of the
special token (camera: $r{=}0$; registers: $r{=}1,\ldots,R$).
The infinitesimal tiebreak $\epsilon_{\text{tb}}$ enforces a fully
deterministic, reproducible eviction order among special tokens (values given in \S\ref{sec:setup}).
This guarantees that global scene anchors are preserved throughout streaming inference.

\vspace{-2mm}
\subsection{Cosine-Similarity-Guided Layer-Wise Budget Allocation}
\label{sec:cosine}
\vspace{-2mm}
A global token budget $B_{\text{total}}$ must be distributed across $L$ transformer
layers.
Uniform allocation $B_l = B_{\text{total}} / L$ ignores the heterogeneous importance
of different layers.

\noindent\textbf{Offline profiling.}
We run inference on representative sequences and register forward hooks
on each global block $\ell$ to capture its input $\mathbf{x}_\ell$ and output
$\mathbf{y}_\ell$.
The mean cosine similarity for layer $\ell$ across frames and sequences is:
\vspace{-4mm}
\begin{equation}
  \bar{\rho}_\ell
    \;=\;
    \frac{1}{|S|}\sum_{s \in S}
      \frac{\langle \mathbf{x}_\ell^{(s)},\, \mathbf{y}_\ell^{(s)} \rangle}
           {\|\mathbf{x}_\ell^{(s)}\|\,\|\mathbf{y}_\ell^{(s)}\|}.
  \label{eq:rho}
\end{equation}
Layers with high $\bar{\rho}_\ell$ are near-identity and hence redundant.

\noindent\textbf{Budget allocation.}
Layer importance is $a_\ell = 1 - \bar{\rho}_\ell$.
The per-layer budget is:
\begin{equation}
  \pi_\ell = \frac{\exp(a_\ell / \tau)}{\sum_{\ell'}\exp(a_{\ell'}/\tau)},
  \qquad
  B_\ell = \bigl\lfloor \pi_\ell \cdot B_{\text{total}} \bigr\rfloor.
  \label{eq:budget}
\end{equation}
Rounding residuals are assigned to the highest-$\pi_\ell$ layer.
The budget vector $\{B_\ell\}$ is computed once offline and fixed at inference time. The full eviction procedure is illustrated in Fig~\ref{fig:similarity}.

\begin{table*}[t]
  \caption{Quantitative comparison on 7-Scenes and NRGBD under different input lengths.
    \textbf{Bold}: best result in each column per input-length block.}
    \vspace{-2mm}
  \label{tab:main_seven_nrgbd}
  \centering
  \small
  \setlength{\tabcolsep}{4pt}
  \resizebox{1\linewidth}{!}{%
  \begin{tabular}{lccccccccccccc}
    \toprule
    \multirow{3}{*}{Method} & \multirow{3}{*}{Input} & \multicolumn{6}{c}{7-Scenes} & \multicolumn{6}{c}{NRGBD} \\
    \cmidrule(lr){3-8}\cmidrule(lr){9-14}
    & & \multicolumn{2}{c}{Acc.$\downarrow$} & \multicolumn{2}{c}{Comp.$\downarrow$} & \multicolumn{2}{c}{NC$\uparrow$}
      & \multicolumn{2}{c}{Acc.$\downarrow$} & \multicolumn{2}{c}{Comp.$\downarrow$} & \multicolumn{2}{c}{NC$\uparrow$} \\
    & & Mean & Med. & Mean & Med. & Mean & Med. & Mean & Med. & Mean & Med. & Mean & Med. \\
    \midrule
    VGGT (\textit{Offline})~\cite{vggt}  & \multirow{7}{*}{\centering 300}
      & OOM & OOM & OOM & OOM & OOM & OOM & OOM & OOM & OOM & OOM & OOM & OOM \\
    StreamVGGT~\cite{streamvggt}         &
      & OOM & OOM & OOM & OOM & OOM & OOM & OOM & OOM & OOM & OOM & OOM & OOM \\
    CUT3R~\cite{cut3r}                   &
      & 0.135 & 0.091 & 0.071 & 0.032 & 0.543 & 0.562
      & 0.224 & 0.126 & 0.074 & 0.012 & 0.579 & 0.624 \\
    Point3R~\cite{point3r}               &
      & 0.047 & 0.027 & 0.029 & 0.011 & 0.563 & 0.596
      & 0.076 & 0.043 & \textbf{0.014} & 0.005 & 0.618 & 0.695 \\
    TTT3R~\cite{ttt3r}                   &
      & 0.041 & 0.025 & 0.024 & 0.005 & 0.565 & 0.599
      & 0.103 & 0.045 & 0.025 & 0.005 & 0.608 & 0.673 \\
    InfiniteVGGT~\cite{infinitevggt}     &
      & 0.040 & 0.015 & 0.025 & 0.005 & 0.570 & 0.607
      & 0.051 & 0.032 & 0.022 & 0.005 & 0.649 & 0.756 \\
    GHOST (ours)                         &
      & \textbf{0.023} & \textbf{0.006} & \textbf{0.022} & \textbf{0.004} & \textbf{0.573} & \textbf{0.612}
      & \textbf{0.046} & \textbf{0.020} & 0.025 & \textbf{0.003} & \textbf{0.690} & \textbf{0.835} \\
    \midrule
    VGGT (\textit{Offline})~\cite{vggt}  & \multirow{7}{*}{\centering 400}
      & OOM & OOM & OOM & OOM & OOM & OOM & OOM & OOM & OOM & OOM & OOM & OOM \\
    StreamVGGT~\cite{streamvggt}         &
      & OOM & OOM & OOM & OOM & OOM & OOM & OOM & OOM & OOM & OOM & OOM & OOM \\
    CUT3R~\cite{cut3r}                   &
      & 0.162 & 0.114 & 0.093 & 0.050 & 0.532 & 0.546
      & 0.315 & 0.215 & 0.101 & 0.032 & 0.551 & 0.572 \\
    Point3R~\cite{point3r}               &
      & 0.049 & 0.023 & 0.026 & 0.009 & 0.559 & 0.589
      & 0.093 & 0.045 & 0.024 & 0.005 & 0.613 & 0.685 \\
    TTT3R~\cite{ttt3r}                   &
      & 0.052 & 0.031 & 0.027 & 0.005 & 0.558 & 0.587
      & 0.140 & 0.070 & 0.058 & 0.014 & 0.599 & 0.657 \\
    InfiniteVGGT~\cite{infinitevggt}     &
      & 0.043 & 0.016 & 0.026 & 0.005 & 0.565 & 0.599
      & 0.069 & 0.040 & 0.034 & 0.005 & 0.653 & 0.763 \\
    GHOST (ours)                         &
      & \textbf{0.026} & \textbf{0.007} & \textbf{0.025} & \textbf{0.004} & \textbf{0.571} & \textbf{0.608}
      & \textbf{0.043} & \textbf{0.020} & \textbf{0.023} & \textbf{0.003} & \textbf{0.682} & \textbf{0.825} \\
    \midrule
    VGGT (\textit{Offline})~\cite{vggt}  & \multirow{7}{*}{\centering 500}
      & OOM & OOM & OOM & OOM & OOM & OOM & OOM & OOM & OOM & OOM & OOM & OOM \\
    StreamVGGT~\cite{streamvggt}         &
      & OOM & OOM & OOM & OOM & OOM & OOM & OOM & OOM & OOM & OOM & OOM & OOM \\
    CUT3R~\cite{cut3r}                   &
      & 0.183 & 0.130 & 0.091 & 0.033 & 0.530 & 0.543
      & 0.326 & 0.243 & 0.132 & 0.042 & 0.556 & 0.582 \\
    Point3R~\cite{point3r}               &
      & 0.063 & 0.026 & 0.031 & 0.015 & 0.555 & 0.583
      & 0.113 & 0.048 & 0.037 & 0.005 & 0.613 & 0.684 \\
    TTT3R~\cite{ttt3r}                   &
      & 0.062 & 0.036 & 0.029 & 0.005 & 0.552 & 0.577
      & 0.165 & 0.084 & 0.095 & 0.015 & 0.594 & 0.648 \\
    InfiniteVGGT~\cite{infinitevggt}     &
      & 0.043 & 0.018 & 0.025 & 0.005 & 0.561 & 0.593
      & 0.080 & 0.054 & 0.037 & 0.008 & 0.643 & 0.746 \\
    GHOST (ours)                         &
      & \textbf{0.027} & \textbf{0.006} & \textbf{0.024} & \textbf{0.003} & \textbf{0.565} & \textbf{0.595}
      & \textbf{0.045} & \textbf{0.021} & \textbf{0.023} & \textbf{0.003} & \textbf{0.672} & \textbf{0.781} \\
    \bottomrule
  \end{tabular}%
  }
  \vspace{-6mm}
\end{table*}

\vspace{-10mm}
\section{Experiments}
\label{sec:experiments}
\vspace{-5mm}
\subsection{Setup}
\label{sec:setup}
\vspace{-2mm}
We evaluate on four benchmarks: Bonn~\citep{bonn} (200--500 frames, depth only), 7-Scenes~\citep{sevenscenes} (300--500 frames), NRGBD~\citep{nrgbd} (300--500 frames),
and Long3D~\citep{infinitevggt} (2{,}128--9{,}545 frames).
Metrics include Accuracy, Completeness, and Normal Consistency (reconstruction), and Abs Rel/$\delta$<$1.25$ (depth).
Baselines are VGGT~\citep{vggt}, StreamVGGT~\citep{streamvggt}, CUT3R~\citep{cut3r}, Point3R~\citep{point3r}, TTT3R~\citep{ttt3r},
and InfiniteVGGT~\citep{infinitevggt} (primary comparison).
Our GHOST uses $B_{\text{total}}{=}1{,}200{,}000$, $\tau{=}0.5$, $\Delta_{\text{boost}}{=}0.3$, $\epsilon_{\text{tb}}{=}10^{-6}$,
$(w_{\text{cam}}, w_{\text{geo}}, w_{\text{temp}}){=}(0.55, 0.55, 0.25)$, $w_f{=}w_k{=}0.5$,
and $(w_{\text{sal}}, w_{\text{dc}}, w_{\text{pc}}){=}(0.28, 0.45, 0.35)$ via grid search on held-out sequences;
cosine-similarity profiling uses 13 sequences on one RTX 4090. All experiments run in anchor-1 mode~\cite{infinitevggt}.

\vspace{-6mm}
\subsection{Main Results}
\label{sec:main}
\vspace{-2mm}

\begin{wraptable}{r}{0.49\textwidth} 
  \vspace{-13mm}
  \centering
  \footnotesize
  \caption{Quantitative comparison on Long3D across five scenes.
    The scene number indicates the sequence length in frames.
    \textbf{Bold}: best result in each column per scene block.}
    \vspace{-2mm}
  \label{tab:main_long3d}
  \small
  \setlength{\tabcolsep}{2pt}
  \resizebox{0.49\textwidth}{!}{
  \begin{tabular}{lp{1.6cm}@{\hspace{1pt}}cccccc}
    \toprule
    \multirow{2}{*}{Method} & \multirow{2}{*}{Scene} 
      & \multicolumn{2}{c}{Acc.$\downarrow$}
      & \multicolumn{2}{c}{NC$\uparrow$}
      & CD$\downarrow$ \\
    \cmidrule(lr){3-4}\cmidrule(lr){5-6}
    & & Mean & Med. & Mean & Med. & \\
    \midrule
    CUT3R~\cite{cut3r}               & \multirow{4}{*}{\makecell{\textit{Classroom} \\ 2128}} 
      & 0.496 & 0.374 & 0.520 & 0.525 & 0.291 \\
    TTT3R~\cite{ttt3r}               &
      & 0.396 & 0.319 & 0.530 & 0.540 & 0.239 \\
    InfiniteVGGT~\cite{infinitevggt} &
      & 0.357 & 0.298 & 0.576 & 0.612 & 0.207 \\
    GHOST (ours)                     &
      & \textbf{0.332} & \textbf{0.277} & \textbf{0.592} & \textbf{0.653} & \textbf{0.197} \\
    \midrule
    CUT3R~\cite{cut3r}               & \multirow{4}{*}{\makecell{\textit{Dormitory} \\ 4208}}
      & 1.800 & 1.372 & 0.501 & 0.495 & 1.102 \\
    TTT3R~\cite{ttt3r}               &
      & 1.965 & 1.749 & 0.515 & 0.509 & 1.147 \\
    InfiniteVGGT~\cite{infinitevggt} &
      & 1.438 & 1.159 & 0.526 & 0.538 & 1.007 \\
    GHOST (ours)                     &
      & \textbf{1.135} & \textbf{0.976} & \textbf{0.541} & \textbf{0.551} & \textbf{0.996} \\
    \midrule
    CUT3R~\cite{cut3r}               & \multirow{4}{*}{\makecell{\textit{Library} \\ 4726}}
      & 1.907 & 1.437 & 0.504 & 0.507 & 1.050 \\
    TTT3R~\cite{ttt3r}               &
      & 2.175 & 1.484 & 0.494 & 0.481 & 1.303 \\
    InfiniteVGGT~\cite{infinitevggt} &
      & 1.121 & 0.821 & 0.508 & 0.514 & 0.846 \\
    GHOST (ours)                     &
      & \textbf{0.745} & \textbf{0.532} & \textbf{0.514} & \textbf{0.533} & \textbf{0.834} \\
    \midrule
    CUT3R~\cite{cut3r}               & \multirow{4}{*}{\makecell{\textit{Badminton} \\ \textit{Court} \\ 6067}}
      & 2.489 & 2.432 & 0.495 & 0.483 & 4.146 \\
    TTT3R~\cite{ttt3r}               &
      & 2.791 & 2.392 & 0.509 & 0.502 & 2.975 \\
    InfiniteVGGT~\cite{infinitevggt} &
      & 1.843 & 1.555 & 0.510 & 0.509 & 1.848 \\
    GHOST (ours)                     &
      & \textbf{1.312} & \textbf{1.007} & \textbf{0.531} & \textbf{0.542} & \textbf{1.612} \\
    \midrule
    CUT3R~\cite{cut3r}               & \multirow{4}{*}{\makecell{\textit{Academic} \\ \textit{Building} \\9545}}
      & 8.062 & 5.650 & 0.496 & 0.491 & 4.638 \\
    TTT3R~\cite{ttt3r}               &
      & 7.710 & 5.793 & \textbf{0.513} & 0.519 & 6.951 \\
    InfiniteVGGT~\cite{infinitevggt} &
      & 5.733 & 4.603 & 0.495 & 0.490 & 3.470 \\
    GHOST (ours)                     &
      & \textbf{4.325} & \textbf{3.961} & 0.511 & \textbf{0.535} & \textbf{3.251} \\
    \bottomrule
  \end{tabular}
  }
  \vspace{-13mm}
\end{wraptable}

Tables~\ref{tab:main_bonn} and~\ref{tab:main_seven_nrgbd} report quantitative results on Bonn, 7-Scenes, and NRGBD under different input lengths.
Table~\ref{tab:main_long3d} reports results on the Long3D benchmark.
GHOST consistently achieves the best performance across nearly all metrics and input
lengths on all four benchmarks.

\noindent\textbf{Bonn.}
In Table~\ref{tab:main_bonn}, GHOST reduces Abs Rel by up to $14.3\%$ and improves $\delta$$<$1.25 over InfiniteVGGT
across all input lengths, confirming the strategy transfers well to depth estimation.

\noindent\textbf{7-Scenes.}
In Table~\ref{tab:main_seven_nrgbd}, at 300 frames, GHOST reduces mean accuracy by $42.5\%$ (0.040$\rightarrow$0.023)
and improves NC over InfiniteVGGT.
The gains hold at 400 and 500 frames, showing geometry-aware eviction retains
structurally informative tokens as sequences grow.

\noindent\textbf{NRGBD.}
In Table~\ref{tab:main_seven_nrgbd}, while existing methods degrade sharply with sequence length---CUT3R by $+45.5\%$,
TTT3R by $+60.2\%$, InfiniteVGGT by $+56.9\%$ from 300 to 500 frames, GHOST
remains stable, with accuracy mean staying around 0.045 across all lengths.

\begin{figure}[t]
  \centering
  \includegraphics[width=\linewidth]{ 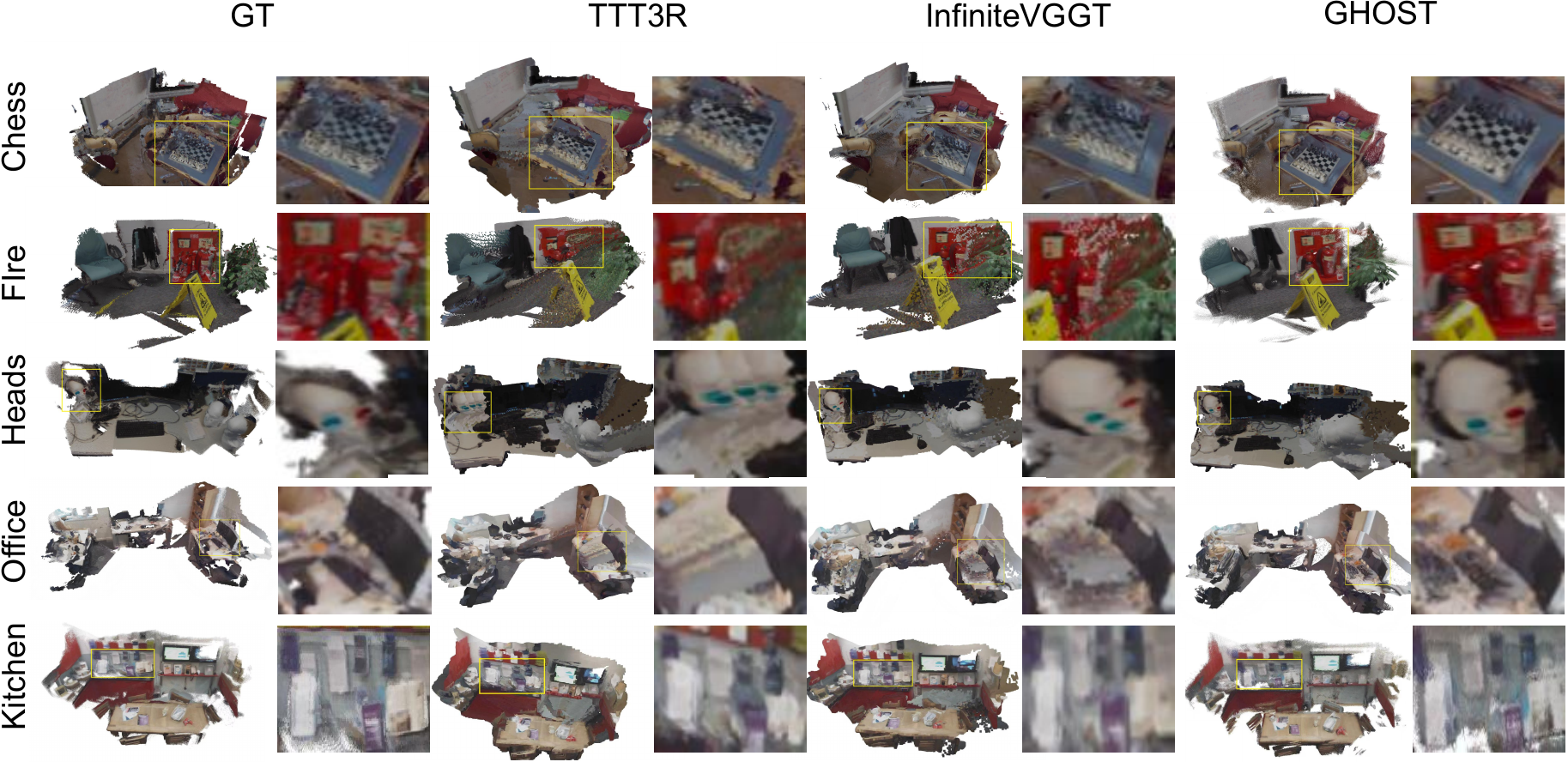}
  \vspace{-3mm}
  \caption{Qualitative reconstruction comparison on 7-Scenes (Chess, Fire, Heads, Office, Kitchen sequences). Columns show GT, TTT3R, InfiniteVGGT, and GHOST results. Yellow boxes highlight fine geometric details where GHOST most closely matches GT.}
  \label{fig:7scenes_recon}
  \vspace{-6mm}
\end{figure}

\noindent\textbf{Long3D.}
On this long-sequence benchmark in Table~\ref{tab:main_long3d},
GHOST achieves the best results on nearly every metric.
Accuracy gains scale with length: $33.5\%$ on Library and $24.6\%$ on Academic Building,
confirming geometry-aware eviction scales gracefully where key-similarity fails.

\vspace{-2mm}
\subsection{Ablation Study}
\label{sec:ablation}
\vspace{-2mm}

We ablate three key design choices of GHOST on 7-Scenes:
the layer-wise budget allocation strategy, the special-token privilege mechanism and the importance weight sensitivity.

\begin{wraptable}{t}{0.6\textwidth} 
    \vspace{-4mm}
  \caption{Ablation of layer budget allocation strategies on 7-Scenes (300 frames).
  \vspace{-2mm}
    \textbf{Bold}: best result.}
  \label{tab:ablation_allocation}
  \centering
  \small
  \setlength{\tabcolsep}{5pt}
  \resizebox{0.6\textwidth}{!}{
  \begin{tabular}{lcccc}
    \toprule
    Scoring function & Acc$\downarrow$ & Comp$\downarrow$ & NC$\uparrow$ & NC Med$\uparrow$ \\
    \midrule
    Uniform allocation                 & 0.027 & 0.023 & 0.565 & 0.608 \\
    Fisher information~\cite{kwon2022fast}                 & 0.033 & 0.024 & 0.570 & 0.610 \\
    Gradient magnitude~\cite{huang2026gradpruner}                 & 0.028 & 0.023 & 0.568 & 0.611 \\
    Hessian saliency~\cite{yang2023global}               & 0.026 & 0.022 & 0.566 & 0.608 \\
    Magnitude pruning~\cite{zhang2024magnitude}                & 0.025 & 0.023 & 0.571 & 0.606 \\
    \midrule
    \textbf{Cosine similarity (GHOST)} & \textbf{0.023} & \textbf{0.022} & \textbf{0.573} & \textbf{0.612} \\
    \bottomrule
  \end{tabular}}
  \vspace{-5mm}
\end{wraptable}

\noindent\textbf{Layer-wise budget allocation strategy.}
Table~\ref{tab:ablation_allocation} compares different layer profiling
strategies with GHOST's scheme.
We evaluate a Uniform allocation baseline (evenly allocating the total budget) alongside various network profiling criteria applied to layer-wise budgeting: Fisher information~\cite{kwon2022fast}, gradient magnitude~\cite{huang2026gradpruner}, Hessian-based salienc~\cite{yang2023global}, and magnitude pruning~\cite{zhang2024magnitude}.
Uniform allocation and all other profiling alternatives underperform the cosine-similarity criterion on Accuracy, confirming that measuring the transformation strength of each layer provides the most reliable budget distribution signal.
The cosine-based score is also the most computationally efficient, requiring no backward passes. 
\begin{wraptable}{t}{0.6\textwidth}
\vspace{-5mm}
  \caption{Effect of special-token privilege on 7-Scenes.}
  \label{tab:ablation_special}
  \vspace{-2mm}
  \centering
  \small
  \setlength{\tabcolsep}{5pt}
  \resizebox{0.6\textwidth}{!}{
  \begin{tabular}{lcccc}
    \toprule
    Configuration & Acc$\downarrow$ & Comp$\downarrow$ & NC$\uparrow$ & NC Med$\uparrow$ \\
    \midrule
    w/o special-token boost ($\Delta_{\text{boost}}{=}0$) & 0.026 & 0.025 & 0.566 & 0.599 \\
    \midrule
    GHOST (w/ boost, $\Delta_{\text{boost}}{=}0.3$)   & \textbf{0.023} & \textbf{0.022} & \textbf{0.573} & \textbf{0.612} \\
    \bottomrule
  \end{tabular}
  }
\vspace{-3mm}
\end{wraptable}

\noindent\textbf{Special-token privilege.}
Table~\ref{tab:ablation_special} shows the effect of disabling the special-token
boost, which allows camera and register tokens to
compete with patch tokens for the shared budget and be evicted.
Without the boost, Accuracy degrades  from 0.023 to 0.026, Comp degrades from 0.022 to 0.025
and NC drops from 0.573 to 0.566, confirming that camera and register tokens encode
structural context that is critical for consistent reconstruction and must be
protected from eviction.

\noindent\textbf{Importance weight sensitivity.}
\label{sec:sensitivity}
GHOST uses six importance weights:
$(w_{\text{cam}}, w_{\text{geo}}, w_{\text{temp}}) = (0.55, 0.55, 0.25)$ for
frame-level scoring and
$(w_{\text{sal}}, w_{\text{dc}}, w_{\text{pc}}) = (0.28, 0.45, 0.35)$ for
token-level scoring.
These were found by a lightweight grid search on five held-out sequences and kept
fixed for all reported experiments.
To verify that GHOST is not sensitive to the precise weight values, we perturb
each group by $\pm 0.1$ around the selected values (one group at a time, keeping
the other fixed) and measure the change in Accuracy Mean on 7-Scenes (300 frames).

Table~\ref{tab:sensitivity} reports results under representative perturbations.
Across all six independent perturbations, Accuracy Mean remains extremely stable, shifting from the optimal 0.023 to at most 0.025 (e.g., when perturbing $w_{\text{dc}}$ by $-0.1$). This minimal degradation
confirms that GHOST is robust to moderate deviations from the tuned weights and
that the grid search result does not overfit to the held-out set.

\begin{table*}[t]
  \centering
  \caption{Importance weight sensitivity on 7-Scenes (300 frames, Acc Mean$\downarrow$).
  Perturbations of $\pm 0.1$ applied to each weight group independently.}
  \vspace{-2mm}
  \label{tab:sensitivity}
  \footnotesize
  \setlength{\tabcolsep}{0pt}
  \renewcommand{\arraystretch}{1.15}
\resizebox{1\textwidth}{!}{
  \begin{tabular*}{\textwidth}{@{\extracolsep{\fill}}lccccccc@{}}
    \toprule
    {Group}
    & \multicolumn{3}{c}{{Frame weights}}
    & \multicolumn{3}{c}{{Token weights}}
    & {GHOST} \\
    \cmidrule(lr){2-4}
    \cmidrule(lr){5-7}
    \cmidrule(lr){8-8}

    {Perturbation}
    & $+0.1$ on $w_{\text{cam}}$
    & $-0.1$ on $w_{\text{geo}}$
    & $+0.1$ on $w_{\text{temp}}$
    & $+0.1$ on $w_{\text{sal}}$
    & $-0.1$ on $w_{\text{dc}}$
    & $+0.1$ on $w_{\text{pc}}$
    & Default \\

    \midrule

    {Acc Mean$\downarrow$}
    & 0.023
    & 0.024
    & 0.024
    & 0.024
    & 0.025
    & 0.023
    & \textbf{0.023} \\

    \bottomrule
  \end{tabular*}
  }
  \vspace{-4mm}
\end{table*}

\vspace{-4mm}
\subsection{Layer Budget Analysis}
\label{sec:layer_budget}
\begin{figure}[t]
  \centering
  \includegraphics[width=0.95\textwidth]{ 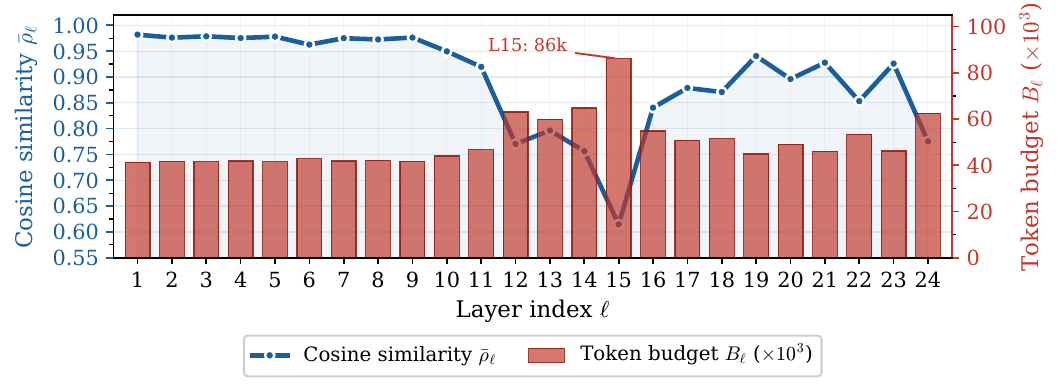}
  \vspace{-3mm}
  \caption{Per-layer cosine similarity $\bar{\rho}_\ell$ (\textcolor{blue}{blue}) and GHOST budget
    $B_\ell$ (\textcolor{orange}{orange}, $\tau{=}0.5$). Lower similarity layers receive larger budgets.}
  \label{fig:budget}
  \vspace{-7mm}
\end{figure}

\vspace{-2mm}
Figure~\ref{fig:budget} shows the cosine-similarity profile of 24 transformer layers
from offline profiling on 7-Scenes.
Early layers (1--9) show high similarity ($\bar{\rho} \approx 0.95$--$1.00$) and receive
smaller budgets, while the 15th layer with the lowest similarity ($\bar{\rho} \approx 0.61$)
gets the largest budget ($86{,}000$).
This trend indicates that layers with stronger input--output transformations require
larger token budgets to preserve informative representations, whereas near-identity
layers can operate effectively with substantially fewer tokens.

\noindent\textbf{Budget sensitivity.}
Table~\ref{tab:budget} reports GHOST on 7-Scenes (300 frames) under varying budgets.
GHOST at $B{=}800{,}000$ surpasses InfiniteVGGT at $B{=}1{,}200{,}000$, and at
$B{=}600{,}000$ (half budget) remains competitive (Acc 0.041 vs.\ 0.040).
At $B{=}600{,}000$, GHOST's KV cache is \textbf{3.51\,GB} vs.\ \textbf{6.94\,GB}
($49\%$ reduction) with a \textbf{$1.75\times$} speedup.
These results demonstrate that GHOST achieves a substantially better
accuracy--efficiency trade-off by dynamically allocating computation to the
most transformation-sensitive layers while aggressively reducing redundant tokens
in near-identity layers.

\vspace{-5mm}
\begin{table}[h]
  \caption{Budget sensitivity on 7-Scenes (300 frames). GHOST achieves comparable
    accuracy at half the budget with significantly lower memory and higher throughput.}
    \vspace{1mm}
  \label{tab:budget}
  \centering
  \small
  \setlength{\tabcolsep}{5pt}
  \resizebox{0.9\textwidth}{!}{
  \begin{tabular}{lrcccccc}
    \toprule
    Method & Budget & Acc$\downarrow$ & Comp$\downarrow$ & NC$\uparrow$ & NC Med$\uparrow$ & KV Cache (GB)$\downarrow$ & Speedup$\uparrow$ \\
    \midrule
    InfiniteVGGT~\cite{infinitevggt} & 1,200,000 & 0.040 & 0.025 & 0.570 & 0.607 & 6.94 & $1.00\times$ \\
    \midrule
    GHOST & 1,000,000 & \textbf{0.023} & \textbf{0.022} & \textbf{0.573} & \textbf{0.613} & 5.80 & $1.38\times$ \\
    GHOST & 800,000   & 0.024 & 0.022 & 0.573 & 0.612 & 4.65 & $1.53\times$ \\
    GHOST & 600,000   & 0.041 & 0.029 & 0.568 & 0.604 & \textbf{3.51} & $\mathbf{1.75\times}$ \\
    \bottomrule
  \end{tabular}
  }
  \vspace{-7mm} 
\end{table}

\section{Conclusion}
\label{sec:conclusion}
\vspace{-3mm}
We presented GHOST, a geometry-hierarchical online streaming token eviction framework
for efficient long-sequence 3D reconstruction.
By exploiting the model's own depth, point, and camera pose outputs as importance
signals, GHOST addresses the fundamental limitation of attention-based eviction, which
cannot distinguish historically redundant tokens from geometrically valuable ones.
Three innovations work in concert: hierarchical dual-level importance scoring,
special-token privilege for structural camera and register tokens, and
cosine-similarity-guided layer-wise budget allocation.
Experimental results validate that geometry-aware importance signals provide
a more principled eviction criterion than attention-based and key similarity-based heuristics,
yielding a favorable result across diverse benchmarks and input lengths.

\bibliographystyle{plainnat}
\bibliography{ghost}
\newpage
\appendix

\section{GHOST Eviction Algorithm}
\label{app:algorithm}

\subsection{Online Incremental Computation}
\label{app:online}

Computing full importance from scratch at every eviction step would require
$\mathcal{O}(T^2)$ operations.
GHOST maintains an \emph{importance cache} that stores per-frame raw scores:
\begin{itemize}[leftmargin=1.5em]
  \item On the first eviction, full importance is computed for all candidate frames.
  \item At subsequent evictions, \emph{only the new frame's} importance is computed
        and concatenated to retained-frame scores, reducing per-step cost to
        $\mathcal{O}(N_p)$.
  \item Camera and geometry raw scores are updated lazily from stored metadata
        (depth, confidence, pose), since frame outputs become available only after
        the next forward pass.
\end{itemize}
This incremental scheme ensures eviction overhead is negligible relative to the
transformer forward pass.

\subsection{Full Eviction Procedure}

\begin{algorithm}[H]
\caption{GHOST Online Token Eviction at frame $t$}
\label{alg:ghost}
\begin{algorithmic}[1]
\REQUIRE Metadata $\mathcal{M}{=}\{(\boldsymbol{\pi}_s,d_s,c^d_s,c^p_s)\}_{s<t}$,
         current patch tokens $\mathbf{F}_t$, importance cache $\mathcal{C}$,
         per-layer budgets $\{B_\ell\}$, KV cache $\mathcal{K}$.
\ENSURE  Updated KV cache $\mathcal{K}'$ with $|\mathcal{K}'_\ell|\le B_\ell$.
\FOR{each layer $\ell = 1, \ldots, L$}
  \IF{$|\mathcal{K}_\ell| \leq B_\ell$}
    \STATE \textbf{continue}
  \ENDIF
  \IF{cached scores $\mathcal{C}_\ell$ available from prior step}
    \STATE Compute $\phi(t,\cdot)$ for current frame only via Eqs.\,\eqref{eq:cam}--\eqref{eq:combined}
    \STATE $\Phi \leftarrow [\mathcal{C}_\ell;\; \phi(t,\cdot)]$
  \ELSE
    \STATE Compute full $\Phi = \{\phi(s,p)\}$ for all candidate frames
  \ENDIF
  \STATE Apply special-token boost via Eq.\,\eqref{eq:boost}
  \STATE $\mathrm{idx} \leftarrow \operatorname{TopK}(\Phi,\; B_\ell)$
  \STATE $\mathcal{K}'_\ell \leftarrow \mathcal{K}_\ell[\mathrm{idx}]$; \quad
         $\mathcal{C}_\ell \leftarrow \Phi[\mathrm{idx}]$
\ENDFOR
\STATE Append current frame KV pairs to each $\mathcal{K}'_\ell$.
\RETURN $\mathcal{K}'$, updated $\mathcal{C}$
\end{algorithmic}
\end{algorithm}

\section{Extended Ablation: Dual-Level Decomposition}
\label{app:ablation_decomposition}

The main paper ablates the importance estimation strategy and the special-token
privilege mechanism.
Here we provide a finer decomposition ablation to justify the hierarchical
dual-level design of the importance score $\phi(t,p)$ (Eq.~\eqref{eq:combined}).

\noindent\textbf{Frame-level vs.\ token-level vs.\ combined.}
Table~\ref{tab:app_decomposition} isolates the contribution of each level by
replacing the full score with its frame-only or token-only counterpart.
Using frame-level importance alone ($\phi = s_{\text{frame}}$) assigns a uniform
importance to all patches within a frame, ignoring within-frame spatial variation
(Accuracy degrades from 0.023 to 0.029, and Normal Consistency drops from 0.573 to 0.568).
Using token-level importance alone ($\phi = s_{\text{token}}$) captures patch
saliency but disregards whether the parent frame is geometrically distinctive
(Accuracy degrades further to 0.033, and Normal Consistency to 0.561).
GHOST combines both, consistently outperforming either ablated variant across all four metrics, confirming
that the two levels are complementary and jointly necessary.

\begin{table}[h]
  \caption{Ablation of hierarchical dual-level decomposition on 7-Scenes (300 frames).
    \textbf{Bold}: best result.}
  \label{tab:app_decomposition}
  \centering
  \small
  \setlength{\tabcolsep}{8pt}
  \begin{tabular}{lcccc}
    \toprule
    Configuration & Acc$\downarrow$ & Comp$\downarrow$ & NC$\uparrow$ & NC Med$\uparrow$ \\
    \midrule
    Frame-level only ($\phi = s_{\text{frame}}$)   & 0.029&0.024 &0.568 &0.607 \\
    Token-level only ($\phi = s_{\text{token}}$)   & 0.033&0.028 &0.561 &0.598 \\
    \midrule
    GHOST (frame + token)                          & \textbf{0.023} & \textbf{0.022} & \textbf{0.573} & \textbf{0.612} \\
    \bottomrule
  \end{tabular}
\end{table}

\noindent\textbf{Frame-level sub-component ablation.}
Table~\ref{tab:app_frame_components} ablates the three sub-signals of the
frame-level score individually: camera pose change $s_{\text{cam}}$, depth
gradient variance $s_{\text{geo}}$, and temporal recency $s_{\text{temp}}$.
Each is removed in turn (weight set to zero, remaining weights renormalised)
to isolate its marginal contribution.
Removing camera pose change ($s_{\text{cam}}$) or temporal recency ($s_{\text{temp}}$) causes Accuracy to worsen from 0.023 to 0.026 and 0.028 respectively. Removing depth gradient variance ($s_{\text{geo}}$) yields the most severe drop, degrading Accuracy to 0.030 and Normal Consistency from 0.573 to 0.558, indicating that structural variation is the strongest single indicator of a frame's long-term geometric value.

\begin{table}[h]
  \caption{Ablation of frame-level importance sub-components on 7-Scenes (300 frames).
    \textbf{Bold}: best result.}
  \label{tab:app_frame_components}
  \centering
  \small
  \setlength{\tabcolsep}{8pt}
  \begin{tabular}{lcccc}
    \toprule
    Configuration & Acc$\downarrow$ & Comp$\downarrow$ & NC$\uparrow$ & NC Med$\uparrow$ \\
    \midrule
    w/o $s_{\text{cam}}$  (no pose change)         & 0.026&0.025 &0.569 &0.605 \\
    w/o $s_{\text{geo}}$  (no depth variance)       & 0.030&0.026 &0.558 &0.595 \\
    w/o $s_{\text{temp}}$ (no recency)              & 0.028& 0.025& 0.566&0.596 \\
    \midrule
    GHOST (all three)                               & \textbf{0.023} & \textbf{0.022} & \textbf{0.573} & \textbf{0.612} \\
    \bottomrule
  \end{tabular}
\end{table}

\section{Extended Ablation: Temperature Sensitivity}
\label{app:ablation_temperature}

The layer-wise budget allocation in GHOST is governed by the temperature
parameter $\tau$ (Eq.~\eqref{eq:budget}), which controls how aggressively the
total budget is redistributed across layers according to their cosine-similarity
profiles.
A higher $\tau$ produces a more uniform distribution; a lower $\tau$ concentrates
budget more heavily on early, high-transformation layers.

Table~\ref{tab:app_temperature} reports GHOST performance on 7-Scenes (300 frames)
under four values of $\tau$.
Performance is remarkably robust across a wide range ($\tau \in [0.3, 0.7]$), with Accuracy fluctuating only marginally between 0.023 and 0.025, and Normal Consistency between 0.571 and 0.573. This confirms
that the offline profiling result generalises and the method is not sensitive to
the precise choice of $\tau$.
We use $\tau = 0.5$ in all main-paper experiments.

\begin{table}[h]
  \caption{Sensitivity of GHOST to temperature $\tau$ on 7-Scenes (300 frames).
    \textbf{Bold}: best result.}
  \label{tab:app_temperature}
  \centering
  \small
  \setlength{\tabcolsep}{8pt}
  \begin{tabular}{ccccc}
    \toprule
    $\tau$ & Acc$\downarrow$ & Comp$\downarrow$ & NC$\uparrow$ & NC Med$\uparrow$ \\
    \midrule
    0.3 &0.024 & 0.022& 0.572& 0.610\\
    \textbf{0.5 (ours)} & \textbf{0.023} & \textbf{0.022} & \textbf{0.573} & \textbf{0.612} \\
    0.7 &0.025 &0.023 & 0.571& 0.608\\
    1.0 & 0.025&0.024 &0.572
    &0.610 \\
    
    \bottomrule
  \end{tabular}
\end{table}

\section{Comparison with LLM KV Cache Eviction Methods}
\label{app:llm_comparison}

A natural question is whether general-purpose KV cache eviction methods
developed for large language models (LLMs) can be applied to streaming 3D
reconstruction.
Table~\ref{tab:app_llm} summarises the key differences between GHOST and
representative LLM KV cache methods.

\begin{table*}[h]
  \caption{Conceptual comparison between GHOST and LLM KV cache eviction methods.}
  \label{tab:app_llm}
  \centering
  \small
  \setlength{\tabcolsep}{5pt}
  \begin{tabular}{lcccc}
    \toprule
    Method & Domain & Importance Signal & Geometry-Aware & Layer-Wise Budget \\
    \midrule
    H2O~\cite{h2o}               & LLM  & Attention score accumulation  & \ding{55} & \ding{55} \\
    StreamingLLM~\cite{streamingllm} & LLM & Fixed sink + recent tokens & \ding{55} & \ding{55} \\
    SnapKV~\cite{snapkv}            & LLM  & Observation window attention  & \ding{55} & \ding{55} \\
    PyramidKV~\cite{pyramidkv}     & LLM  & Layer-wise attention budget   & \ding{55} & \ding{51} \\
    InfiniteVGGT~\cite{infinitevggt}      & 3D   & Key-query cosine similarity   & \ding{55} & \ding{55} \\
    \midrule
    \textbf{GHOST (ours)}                 & 3D   & Geometry-hierarchical score   & \ding{51} & \ding{51} \\
    \bottomrule
  \end{tabular}
\end{table*}

LLM KV cache methods rely exclusively on attention-based signals (accumulated
attention scores, recent token windows, or observation-window attention patterns)
to decide which tokens to evict.
These signals are query-dependent and context-agnostic: they measure how much the
\emph{current} query attends to past tokens, which correlates poorly with the
long-term geometric value of historical 3D observations
(Spearman $\rho = -0.07$ with pose change, $\rho = -0.31$ with depth gradient
variance, as shown in the main paper).

GHOST differs in two fundamental respects.
First, the importance signal is \emph{geometry-grounded}: it is derived from the
model's own 3D outputs (depth maps, point confidence, camera poses) rather than
from attention patterns.
This makes the score query-independent and reflective of a token's long-term
structural value.
Second, GHOST introduces \emph{geometry-aware layer-wise budget allocation}
guided by offline cosine-similarity profiling, which concentrates the token
budget where transformer layers perform the strongest transformations.
PyramidKV~\cite{pyramidkv} also allocates budgets layer-wise, but based
on attention entropy rather than transformation strength, and does not exploit any
domain-specific structure.
These design differences make direct application of LLM KV cache methods to
streaming 3D reconstruction ineffective, and motivate the geometry-specialised
design of GHOST.

\section{Broader Applicability of GHOST}
\label{app:applicability}

GHOST is presented as a drop-in replacement for the InfiniteVGGT eviction module,
but the underlying framework is designed to be broadly applicable to any
streaming transformer that produces geometry-relevant outputs at each frame.
We discuss the prerequisites for transferring GHOST to other architectures and
identify promising directions for future work.

\noindent\textbf{Prerequisites.}
GHOST's frame-level importance score $s_{\text{frame}}(t)$ requires three signals:
(i) a camera pose estimate $\boldsymbol{\pi}_t$ (translation, rotation, focal
length), (ii) a per-frame depth map $d_t$, and (iii) per-patch confidence maps
$c^d_t, c^p_t$ for depth and point reconstruction quality.
These outputs are already produced natively by InfiniteVGGT (and its parent
architecture StreamVGGT) at every forward pass, incurring no additional overhead.

The token-level importance score $s_{\text{token}}(t,p)$ requires patch-level
depth and point confidence maps, which are straightforwardly available from any
model with dedicated depth or point-map heads.
The visual saliency signal $s_{\text{sal}}(t,p)$ is computed from the patch
feature map and requires only a single gradient operation, making it architecture-agnostic.

\noindent\textbf{Candidate architectures.}
Several recent streaming 3D reconstruction models satisfy these prerequisites:

\begin{itemize}[leftmargin=*, itemsep=2pt]
  \item \textbf{MASt3R-SfM}~\cite{mast3r}: produces per-pair depth and confidence
    maps alongside dense feature descriptors; the depth and confidence outputs
    directly provide the signals required for GHOST scoring.
  \item \textbf{MonST3R}: a monocular streaming variant of DUSt3R that outputs
    depth maps with uncertainty estimates per frame, satisfying the depth and
    confidence requirements.
  \item \textbf{CUT3R}~\cite{cut3r}: maintains a recurrent state over past frames
    and outputs depth and point maps at each step; the recurrent state could be
    compressed using GHOST's geometry-hierarchical importance scores.
\end{itemize}

\noindent\textbf{Proxy signals for models without geometry heads.}
For streaming transformers that do not natively output depth or confidence maps
(e.g., pure feature-matching models), proxy signals can be substituted: optical
flow magnitude can approximate camera pose change, image gradient variance can
approximate depth gradient variance, and attention entropy can serve as a
proxy for patch-level confidence.
These substitutions degrade the quality of the importance signal but preserve
the hierarchical dual-level structure of GHOST.

\noindent\textbf{Limitations.}
GHOST is not directly applicable to architectures that lack any causal structure
(e.g., fully offline joint-attention models such as VGGT), or to models where
the KV cache is not the primary memory bottleneck.
Extending GHOST to dynamic scenes (moving objects, non-rigid deformation) remains
an open challenge, as the current frame-level score assumes a static scene and
may over-retain frames from geometrically similar but dynamically changed regions.

\begin{figure}[t]
  \centering
  \includegraphics[width=0.85\linewidth]{ 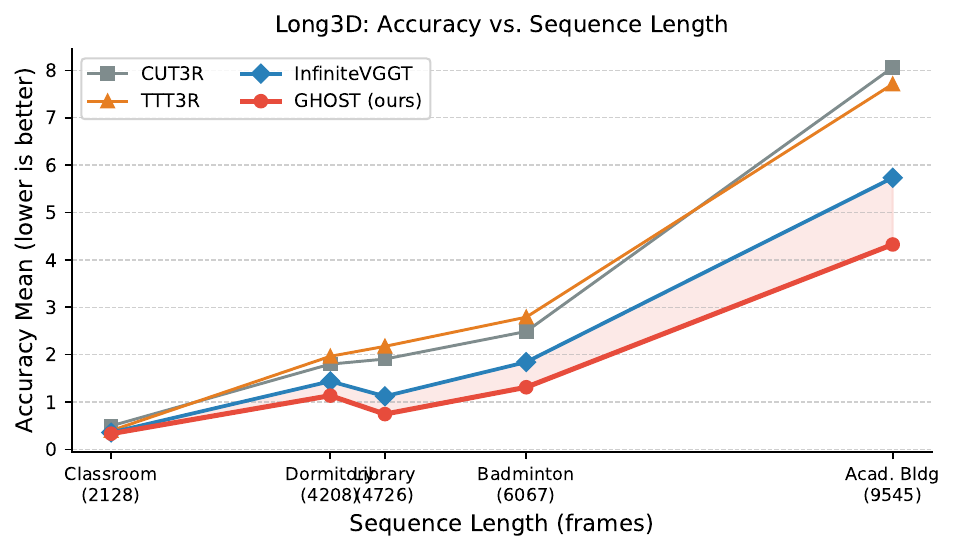}
  \caption{Accuracy Mean ($\downarrow$) versus sequence length on the Long3D
    benchmark.
    The shaded region highlights the gap between GHOST and InfiniteVGGT.
    GHOST's advantage over key-similarity eviction (InfiniteVGGT) grows with
    sequence length, confirming that geometry-aware eviction scales more
    gracefully to very long sequences.}
  \label{fig:app_scalability}
\end{figure}
\section{Long-Sequence Scalability Analysis}
\label{app:scalability}

A key motivation for GHOST is that key-similarity eviction degrades
disproportionately as sequences grow longer, because the current query is
increasingly remote from early frames, causing them to receive systematically
low similarity scores and be evicted even when they contain unique scene content.
We verify this hypothesis using the Long3D benchmark, which spans sequences
from 2{,}128 to 9{,}545 frames---an order of magnitude longer than the
7-Scenes and NRGBD sequences in the main paper.

Figure~\ref{fig:app_scalability} plots Accuracy Mean against sequence length
for all methods.
Two trends are evident.
First, all methods degrade as sequence length increases, reflecting the growing
difficulty of retaining a fixed-budget representation of an increasingly long
history.
Second, the \emph{gap} between GHOST and InfiniteVGGT widens with sequence
length: at 2{,}128 frames (Classroom), GHOST improves Accuracy by $7.0\%$
(0.357$\rightarrow$0.332); at 4{,}726 frames (Library), the improvement reaches
$33.5\%$ (1.121$\rightarrow$0.745); and at 9{,}545 frames (Academic Building),
$24.6\%$ (5.733$\rightarrow$4.325).
This superlinear scaling of the benefit confirms the core hypothesis: the longer
the sequence, the more severely key-similarity eviction misfires by discarding
geometrically distinctive early frames, and the more geometry-aware eviction pays
off.

CUT3R and TTT3R exhibit particularly severe degradation at extreme sequence
lengths (Academic Building: 8.062 and 7.710 respectively), as their sequential
update strategies accumulate drift without a principled mechanism to preserve
globally informative historical observations.
GHOST's geometry-hierarchical scoring explicitly accounts for viewpoint coverage
and depth diversity across the full sequence history, yielding substantially more
graceful degradation.
\end{document}